\documentclass[conference]{IEEEtran}
\usepackage{times}

\usepackage{graphics} 
\usepackage{amsmath} 
\usepackage{amssymb}  
\usepackage{siunitx}
\usepackage{mathtools}
\usepackage{amsthm}
\usepackage{booktabs}
\usepackage{multirow}
\usepackage{bbm}
\usepackage[dvipsnames]{xcolor}
\usepackage{hyperref}
\usepackage{subcaption} 
\usepackage{enumerate}
\usepackage{mathrsfs}

\usepackage[bottom=57pt,top=54pt, left=54pt, right=54pt]{geometry}   
\usepackage{graphicx}
\usepackage{adjustbox}
\usepackage{pbox}
\usepackage{dblfloatfix}
\usepackage{hyperref}
\usepackage{caption}
\usepackage[nospace]{cite}

\usepackage[ruled]{algorithm2e} 
\usepackage{xcolor} 
\usepackage{tikz}
\usetikzlibrary{fit,calc}

\colorlet{mypink}{red!40}
\colorlet{myblue}{cyan!60}

\usepackage{array,multirow}
\usepackage{makecell}
\usepackage[T1]{fontenc}
\usepackage{microtype}
\usepackage{tikz}
\usetikzlibrary{positioning}
\usepackage{pgfplots}
\usepackage{xspace}
\usepackage[numbers]{natbib}
\usepackage{multicol}

\usepackage{amssymb}
\usepackage{pifont}
\newcommand{\etal}{\emph{et~al.}}

\newcommand{\parag}[1]{\vspace{-4px} \vskip8pt \noindent \textbf{#1} \hspace{3px}}

\usepackage{soul}

 \newcommand{\removed}[1]{}
\let\vec\mathbf

\DeclareMathOperator*{\argmin}{arg\,min}
\newcommand{\oursubsubsection}[1]{\par\textbf{#1}:}

\begin{document}

\title{
DexEvolve: Evolutionary Optimization for Robust and Diverse Dexterous Grasp Synthesis}

\author{\authorblockN{René Zurbrügg, Andrei Cramariuc, Marco Hutter} \authorblockA{ETH Zürich}}



%

\maketitle

\begin{abstract}
Dexterous grasping is fundamental to robotics, yet data-driven grasp prediction heavily relies on large, diverse datasets that are costly to generate and typically limited to a narrow set of gripper morphologies. Analytical grasp synthesis can be used to scale data collection, but necessary simplifying assumptions often yield physically infeasible grasps that need to be filtered in high-fidelity simulators, significantly reducing the total number of grasps and their diversity.

We propose a scalable generate-and-refine pipeline for synthesizing large-scale, diverse, and physically feasible grasps. Instead of using high-fidelity simulators solely for verification and filtering, we leverage them as an optimization stage that continuously improves grasp quality without discarding precomputed candidates. More specifically, we initialize an evolutionary search with a seed set of analytically generated, potentially suboptimal grasps. We then refine these proposals directly in a high-fidelity simulator (Isaac Sim) using an asynchronous, gradient-free evolutionary algorithm, improving stability while maintaining diversity. In addition, this refinement stage can be guided toward human preferences and/or domain-specific quality metrics without requiring a differentiable objective.
We further distill the refined grasp distribution into a diffusion model for robust real-world deployment, and highlight the role of diversity for both effective training and during deployment. Experiments on a        
newly introduced Handles dataset and a DexGraspNet subset demonstrate that our approach achieves over 120 distinct stable grasps per object (a 1.7-6x improvement over unrefined analytical methods) while outperforming diffusion-based alternatives by 46-60\% in unique grasp coverage. \\ \noindent Project page: \url{https://dexevolve.github.io}

\end{abstract}

\IEEEpeerreviewmaketitle

\section{Introduction}

Dexterous grasping is a fundamental capability in robotics, enabling a wide range of applications, including industrial automation, service robotics, and human-robot collaboration. 
Traditional approaches to grasp synthesis primarily rely on analytical methods, which sample contact points on an object’s surface and evaluate their force-closure properties by solving conical optimization programs~\cite{liu2020new}. While analytical optimization can generate diverse grasp candidates, many of these are not dynamically feasible due to simplifying assumptions in the underlying models (e.g., coarse contact dynamics, friction approximations, and simplified collisions).



Two-stage approaches first generate grasp candidates, then filter them using high-fidelity, non-differentiable physics simulators such as Isaac Sim~\cite{isaac-sim} or MuJoCo~\cite{mujoco}. Classical methods like GraspIt!~\cite{graspit} use shape-guided heuristics and quality metrics to generate and evaluate candidates. More recent methods~\cite{wang2023dexgraspnet,turpin2023fastgraspd,liu2021synthesizing,graspqp,chen2025bodex,chen2025dexonomy} relax or approximate traditional force (or form) closure conditions, enabling integration with gradient-based optimization and kinematic simulators. Despite these advances, both paradigms share a fundamental limitation: the separation between generation and validation. Grasps are proposed under simplified or relaxed assumptions, then verified in high-fidelity simulation. This post-selection strategy is inherently sample-inefficient as most proposals are discarded during verification, and those that pass are often still sub-optimal since the generation process cannot leverage direct physical feedback to refine candidates during optimization. 
We argue that, rather than treating high-fidelity simulation as a binary accept/reject filter, grasp proposals can benefit substantially from further refinement via direct optimization within the physics simulator.


Direct simulator-based refinement is challenging for two reasons. First, evaluating grasp stability in a physics simulator is significantly more expensive than kinematic checks. It requires applying interaction forces over a predefined time horizon and verifying that the object remains stable. Second, most high-fidelity simulators are non-differentiable, which prevents the use of conventional gradient-based optimization methods. Hence, the few works that investigate direct optimization in high-fidelity simulators are typically limited to wrist-pose optimization and do not consider the full grasp configuration in order to simplify the search space \cite{huber2024speeding,huber-icra-dr-grasping}.

\begin{figure}[t!]
    \centering
    \vspace{8mm}
    \includegraphics[width=1.0\linewidth]{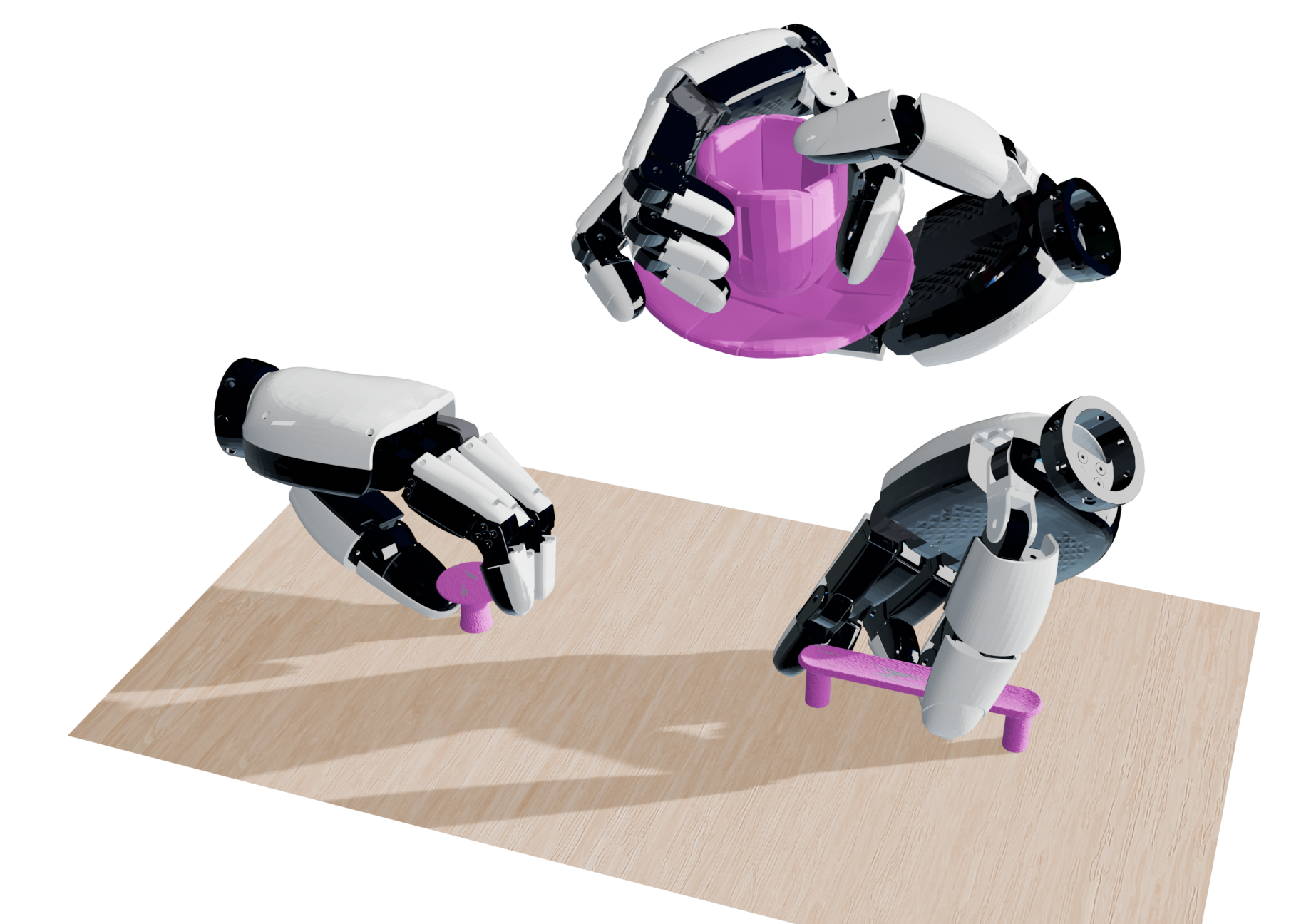}
    \caption{\textbf{Visualization of Generated Grasps using our Evolutionary Refinement} visualized for two \emph{Handles} and one \emph{object} asset. Our evolutionary approach generates diverse, physically stable grasps by refining candidates directly within high-fidelity simulation.}
    \vspace{-7mm}
    \label{fig:teaserFigure}
\end{figure}
To address these shortcomings, we propose a two-stage approach to using high-fidelity simulators for grasp refinement, without differentiability constraints.
We start with a diverse set of analytically generated grasp proposals from~\cite{graspqp}. In practice, our framework does not restrict the initialization strategy, and other methods are equally applicable for obtaining an initial pool of grasp candidates. Subsequently, we refine these proposals directly in a high-fidelity simulator (Isaac Sim) using a gradient-free evolutionary (genetic) algorithm that runs asynchronously. The asynchronicity and parallelization of the process allows many unpromising candidate grasps to be quickly discarded, thus efficiently allocating compute toward promising regions of the grasp space. 

Concretely, we reinterpret high-fidelity simulation not as a binary accept/reject filter, but as a black-box objective that can be optimized directly: instead of discarding the majority of the seed grasps, we iteratively \emph{repair} and \emph{improve} them via evolutionary refinement, yielding a substantially higher number of stable, dynamically feasible grasps (Fig.~\ref{fig:teaserFigure}) while preserving coverage of multiple grasp modes. Our contribution includes the use of density-aware selection and archive-based insertion rules that explicitly suppress redundant grasp clusters and promote exploration of underrepresented regions of the grasp space. Moreover, because our refinement is gradient-free, it naturally supports steering with non-differentiable, domain-specific metrics (e.g., interaction objectives for handles) and even human preference signals.
Finally, we distill the refined grasp distribution into a conditional diffusion model to enable fast inference from partial observations, bridging large-scale dataset generation and deployable grasp prediction.

In summary, our contributions are:

\begin{itemize}
\item \textbf{Generate-and-refine evolutionary grasp synthesis:} A scalable pipeline that uses an asynchronous, gradient-free evolutionary optimizer to refine an initial seed of dexterous grasp configurations directly in a high-fidelity, non-differentiable simulator, using density-aware selection and archive-based insertion to mitigate mode collapse and preserve multimodal grasp diversity.
\item \textbf{Gradient-free guidance:} Our framework allows the refinement step to be guided towards human preference and/or task-specific quality metrics, without requiring the objective to remain differentiable. 
\item \textbf{Real-world deployment:} By training a point-cloud-conditioned diffusion model on the refined dataset, resulting in not only more diverse but real-world applicable grasp predictions from partial point-cloud observations.
\end{itemize}

\section{Related Work}
\subsection{Dexterous Grasping Datasets and Grasp Synthesis}
A wide range of datasets~\cite{casas2024multigrippergraspdatasetroboticgrasping, turpin2023fastgraspd, wang2023dexgraspnet, li2022gendexgrasp, zhang2025graspxl, 9560844} provide synthesized grasp poses for rigid objects across different grippers. These datasets vary in scale, from sets with a few hundred thousand grasps~\cite{liu2020deep, li2022gendexgrasp} to large-scale datasets containing a million or more grasp annotations~\cite{9560844, li2022gendexgrasp, wang2023dexgraspnet, casas2024multigrippergraspdatasetroboticgrasping}. In terms of gripper diversity, many focus on parallel jaw grippers~\cite{9560844,shao2020unigrasp}, while others incorporate dexterous hands such as the Shadow or Allegro Hand~\cite{casas2024multigrippergraspdatasetroboticgrasping, turpin2023fastgraspd, graspqp}. 

These datasets are typically generated through grasp synthesis methods. Traditional sampling-based approaches~\cite{Miller2004, casas2024multigrippergraspdatasetroboticgrasping, Dai2018} explore possible hand poses to optimize grasp quality metrics but are computationally expensive for high-DoF dexterous hands. Recent gradient-based methods~\cite{liu2021synthesizing, li2022gendexgrasp, turpin2022grasp, turpin2023fastgraspd} leverage differentiable grasp quality metrics for faster convergence. However, classical metrics like $Q_1$~\cite{Ferrari1992PlanningOG} are not inherently differentiable, leading to relaxed formulations~\cite{liu2020deep, liu2021synthesizing} that may oversimplify grasp stability by neglecting frictional forces~\cite{liu2021synthesizing, wang2023dexgraspnet}. Other approaches incorporate differentiable kinematic simulations with velocity-based metrics~\cite{turpin2022grasp, turpin2023fastgraspd}.

Our dataset extends this diversity by contributing a novel collection of door handle and knob assets with grasp annotations for the XHand~\cite{robotera_xhand1_2024_biblatex}. By modeling these assets after products available in physical stores, our dataset enables real-world testing that supports reproducible evaluation and sim-to-real transfer across different works. While~\cite{urakami2019doorgym} provides synthetic door knobs for reinforcement learning, but does not include grasp predictions nor corresponding physical assets.

\begin{figure*}
    \centering
    \includegraphics[width=1.0\linewidth]{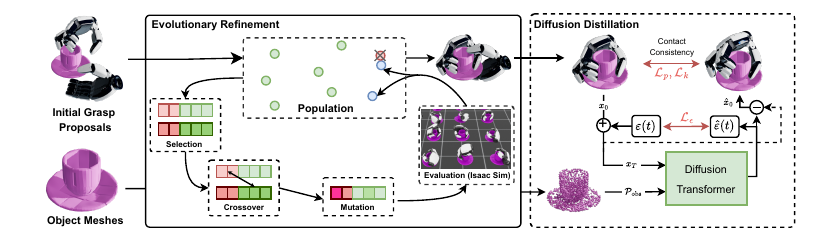}
    \caption{\textbf{Method Overview.} Given object meshes, we first generate an initial set of analytical grasp proposals. We then refine these candidates in a high-fidelity simulator using an asynchronous gradient-free evolutionary loop (selection, crossover, mutation, and physics-based evaluation) to obtain a diverse set of physically feasible grasps. Finally, we distill the refined grasp distribution into a diffusion model conditioned on noisy pointcloud observations, enabling efficient sampling at inference time under partial observability.}
    \label{fig:method_overview}
\end{figure*}

\subsection{Evolutionary Algorithms and Quality Diversity}
Evolutionary algorithms~\cite{sampson1976adaptation} are gradient-free methods that improve a population of solutions via mutation and fitness-based selection. Genetic Algorithms~\cite{ANDROULAKIS1991217} encode solutions as genes and evolve them through \emph{selection}, \emph{crossover}, and \emph{mutation}, and have been used for tasks such as system identification~\cite{179842,bjelonic2025towards} and hyperparameter tuning~\cite{xiao2020efficient}.

In grasp prediction, their use remains limited: Shukla \textit{et al.}~\cite{shukla2021robotic} optimize two-finger gripper poses, Huber \textit{et al.}~\cite{huber2024speeding} optimize wrist placement for diverse hand poses, and Huber \textit{et al.}~\cite{huber2025qdgset} release a large-scale two-finger dataset generated with QDG-6DOF~\cite{huber-icra-dr-grasping}. However, no recent work optimizes full dexterous grasp configurations beyond wrist placement or parallel-jaw setups.

\subsection{Diffusion Models for Grasping}
Recent work has explored diffusion and other generative models for sampling multimodal dexterous grasps conditioned on high-level intent and partial observations, including Transformer-based architectures with pointcloud conditioning~\cite{xu2024dexterous} and approaches that use language as an explicit conditioning signal~\cite{wei2024grasp, lu2024ugg}. Zhong \etal~\cite{zhong2025dexgrasp} target more universal dexterous grasping by incorporating physics-aware modeling into the grasp generation pipeline and serve as the core building block for our diffusion model. Additionally, \cite{zhu2025evolvinggrasp} introduces Direct Policy Optimization for Grasp Pose Diffusion to align grasp generation with human feedback.

A key commonality of these generative approaches is that they typically fit or distill a grasp distribution from existing data rather than synthesizing and validating new grasps through direct optimization in high-fidelity simulators. This distillation enables efficient inference and conditioning on partial observations like point clouds for fast on-hardware deployment. We likewise distill a grasp distribution using an architecture similar to DexGraspAnything~\cite{zhong2025dexgrasp}, but add geometric consistency constraints and diffuse the full wrist pose rather than only position. Unlike pure generative modeling, our framework uses simulator-in-the-loop evolutionary refinement to explicitly improve physical feasibility and stability of grasp configurations, then distills the refined distribution for efficient point-cloud-conditioned inference.

\section{Method}

Given an accurate 3D object model and a dexterous hand, our goal is to synthesize a diverse set of feasible grasp configurations $\mathcal{G}=\{G_i\}_{i=1}^N$ that satisfy force closure, i.e., can resist arbitrary bounded external wrenches by applying admissible contact forces subject to Coulomb friction constraints.
We first generate a diverse seed set with an analytical optimizer (GraspQP~\cite{graspqp}), though any grasp generation method could serve this purpose, then refine these candidates using the evolutionary procedure, detailed in Sec.~\ref{subsec:evolutionary_refinement}.
For real-world deployment, we distill the resulting grasp dataset into a point-cloud-conditioned diffusion model in Sec.~\ref{subsec:diffusion_distillation}.
An overview of the full pipeline is shown in Fig.~\ref{fig:method_overview}.

\subsection{Grasp Representation}
For a given robotic gripper with $n_q$ joints, we parameterize each grasp as $G = ({\chi}, {q}, {\Delta q}_{cmd} )$, where ${\chi} = (\chi_{pos}, \chi_{orient})  \in SE(3)$ denotes the wrist pose, ${q} \in \mathbb{R}^{n_q}$ the joint states of the $n_q$ joints of the robot and ${\Delta q}_{cmd} \in \mathbb{R}^{n_q}$ the desired delta joint commands to apply a gripping force.

\subsection{Evolutionary Refinement}
\label{subsec:evolutionary_refinement}In the following, we outline our evolutionary refinement procedure, which optimizes dexterous grasp candidates directly in Isaac Sim via massively parallel rollouts with early rejection. We introduce the simulator-derived fitness and the genetic operators (density-aware selection, mutation, structured crossover) that improve stability while preserving multimodality. Following evolutionary computation conventions, we refer to each grasp as an individual and to the set of grasp candidates as the population/archive.

Evolutionary refinement in high-dimensional grasp spaces is prone to \emph{mode collapse}, i.e., repeatedly producing minor variants of the same grasp. To maintain diversity, we treat the population as an \emph{archive} and gate insertions using a novelty criterion as done in quality-diversity methods \cite{mouret2015illuminating}: a newly evaluated candidate is inserted only if it is sufficiently far from the current archive under a task-relevant metric $d(\cdot,\cdot)$; otherwise it competes locally with its nearest neighbor. Additionally, to promote diversity during reproduction, we bias the selection process toward lower-density regions.
\oursubsubsection{Insertion of New Grasps}
This step updates the archive by inserting novel candidates and replacing redundant ones only when they improve fitness.
Given a candidate $G$ and the current population $\mathcal{G}=\{G_j\}_{j=1}^{|\mathcal{G}|}$, we compute its nearest neighbor
\begin{equation}
j^\star \;=\; \argmin_{j} d(G,G_j),
\end{equation}
where $d(\cdot,\cdot)$ is a weighted distance metric over position, orientation, and joint configuration:
\begin{equation}
\begin{aligned}
d(G,G')
&= \lambda_\text{pos} \left\lVert G_{\text{pos}} - G'_{\text{pos}} \right\rVert_2^2
 + \lambda_\text{orient} \left\lVert G_{\text{orient}} - G'_{\text{orient}} \right\rVert_2^2 \\
&\quad + \lambda_\text{joints} \left\lVert G_{\text{joints}} - G'_{\text{joints}} \right\rVert_2^2.
\end{aligned}
\label{eq:insertion_distance}
\end{equation}
If $d(G,G_{j^\star})\ge \tau$, the candidate is considered novel and inserted: $\mathcal{G}\leftarrow \mathcal{G}\cup\{G\}$. If $d(G,G_{j^\star})<\tau$, we do not grow the archive. Instead, the candidate competes with its nearest neighbor and replaces it only if it improves fitness:
\begin{equation}
G_{j^\star} \leftarrow G
\quad\text{if}\quad
\mathcal{F}(G) > \mathcal{F}(G_{j^\star}).
\label{eq:replacement_rule}
\end{equation}
This replacement rule preserves diversity by preventing redundant insertions while still allowing local improvements to update an existing individual.

\oursubsubsection{Fitness Function}
The fitness function for the evolutionary algorithm measures the quality of a grasp and is defined as a weighted sum of three components: an environment-lifetime score, a contact-distance penalty, and a penetration penalty.
\begin{equation}
    \mathcal{F}(G) = {E}_{lifetime} -  w_{dis}{E}_{dis} - w_{pen}{E}_{pen},
\end{equation} 
where $w_{dis}$ and $w_{pen}$ denote the respective weights.
$E_{lifetime}$ measures robustness under a fixed disturbance protocol: within each simulation environment, forces are applied sequentially along each canonical axis in both directions ($x, -x, y, -y, z, -z$) over a predefined time horizon, and $E_{lifetime}$ is incremented for each step without failure.
$E_{dis}$ is a weak shaping term that penalizes grasps that remain far from contact-feasible configurations (e.g., poses that lie largely in free space).
$E_{pen}$ penalizes hand--object interpenetration, and we additionally terminate evaluations early if penetration exceeds a predefined threshold.



\oursubsubsection{Selection} We rely on tournament selection~\cite{GOLDBERG199169} to pick parents for crossover. In each tournament, we randomly sample $k$ individuals from the population and select the two highest-scoring candidates.
As discussed at the beginning of Sec.~\ref{subsec:evolutionary_refinement}, selecting purely by raw fitness can accelerate mode collapse~\cite{katoch2021review}. To preserve diversity, we therefore perform tournament selection using a density-aware reweighting of each individual's score $\mathcal{F}'(G_i)$ based on the current population:
\begin{equation}
\label{eq:density-sampling}
    \mathcal{F}_i' = \mathcal{F}(G_i) \cdot \left(\sum_{G_j \in \mathcal{B}_r(G_i)}\left(1 - \frac{d(G_i, G_{j})^p}{r^p}\right)\right)^{-1},
\end{equation}
where $r$, $p$ are tunable parameters that define how strongly clustering is suppressed. In simple terms, if an individual $G_i$ has many similar grasp poses (neighbors) $G_j$ within a ball of radius $r$, the chances of it ($G_i$) being selected will be lower. Note that in the extreme case of (a) having no neighbors ($\mathcal{B}_r(G_i)=\{G_i\}$), the fitness of grasp $G_i$ will remain unchanged ($\mathcal{F}' \rightarrow \mathcal{F}$) and (b) in the case of all grasps being equal the fitness of the grasps will be reduced by $1/{|\mathcal{G}|}$ ($F \rightarrow \frac{F}{{|\mathcal{G}|}}$).
While the selection strategy and the archival-based insertion use the same metric and drive a similar purpose, the selection strategy aims to steer the distribution towards promising samples, whereas the insertion strategy ensures the final population remains diverse once the algorithm terminates.

\oursubsubsection{Mutation}
The mutation operator is applied to the first two parameters (pose and joint states) of the grasp configuration with a probability $p_{\text{mutation}}$.
The mutation step size is drawn from a Gaussian distribution with a zero mean and a standard deviation of 
$
\Sigma = \mathrm{diag}(\Sigma_{{\chi}}, \Sigma_{{q}})
$.
Note that we rely on the Euler angle representation for the wrist pose ${\chi}$, so that mutation can be independently applied to the roll, pitch, and yaw angles.

\oursubsubsection{Crossover}
Genetic algorithms typically use a crossover operator to combine genes from two parents into a new individual. However, naively combining grasp configurations from two parents yields many infeasible grasps.
In line with other research\cite{chen2025dexonomy,graspqp}, we observe that grasps tend to follow different grasp taxonomies with similar finger configurations. 
Furthermore, we observe that many objects can be grasped from various poses with similar finger configurations.  
Hence, we propose exchanging the finger configurations or full grasp pose information as a whole, essentially fixing the cross-over point.
More formally, we randomly select two parent grasps $G_i$ and $G_j$ from the population and generate a new offspring grasp $G_c$ by swapping the finger configurations (joint states) of the two parents with a probability $p_{\text{crossover}}$.

\begin{figure}[t!]
\vspace{-5mm}
    \centering
    \includegraphics[width=1.0\linewidth]{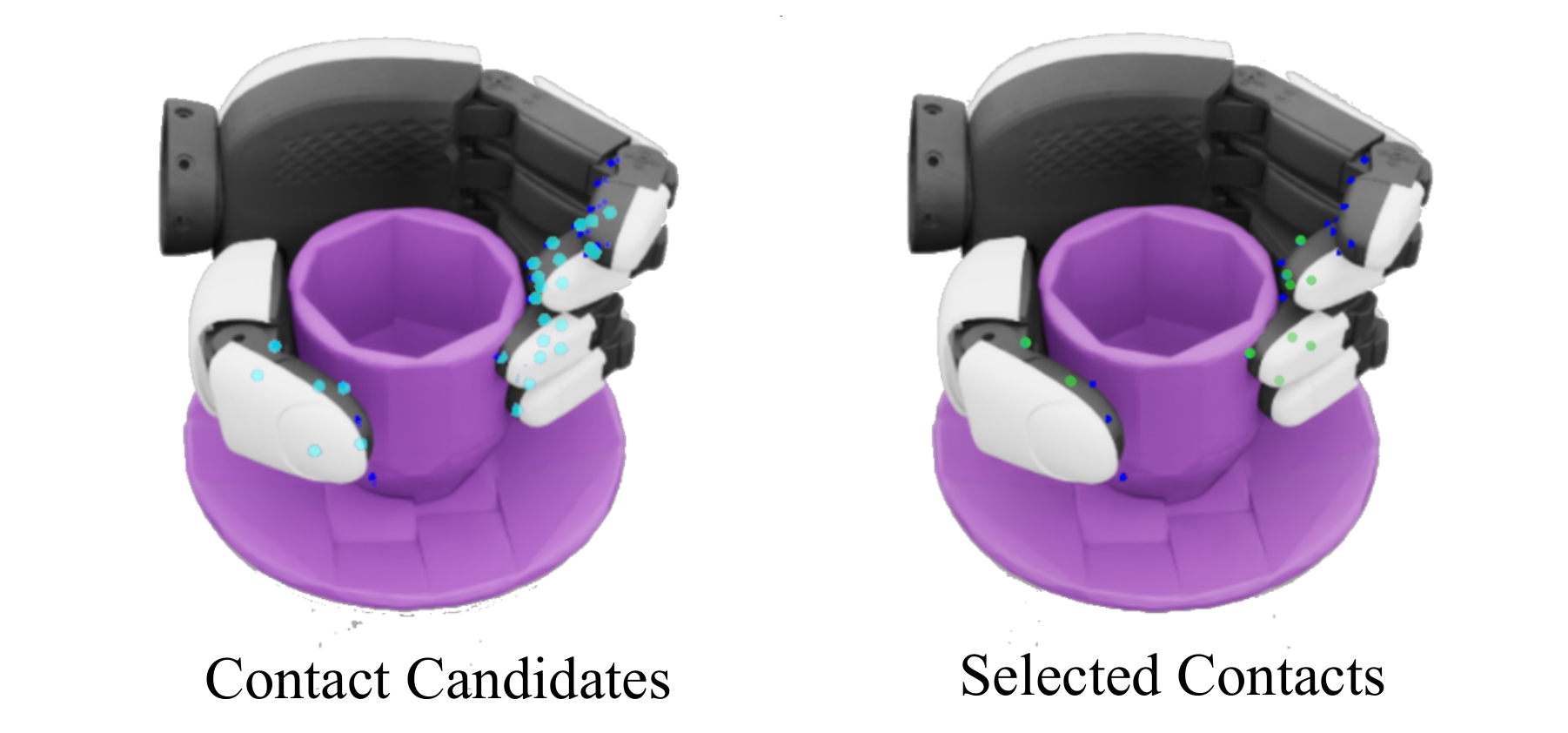}
 \caption{\textbf{Contact Point Selection for Dexterous Grasping.} Two views of a dexterous hand (Xhand) grasping a cup from the Objects dataset. Light blue points indicate candidate contact points on the object surface that lie within the contact distance threshold, while dark blue points denote points outside this threshold. Green points represent the final active contact points selected via farthest point sampling (FPS), which are used to compute grasping commands through the contact Jacobian.}
  \vspace{-5mm}
    \label{fig:hands_overview}
\end{figure}

\begin{figure*}[t]
    \centering
    \includegraphics[width=0.8\linewidth]{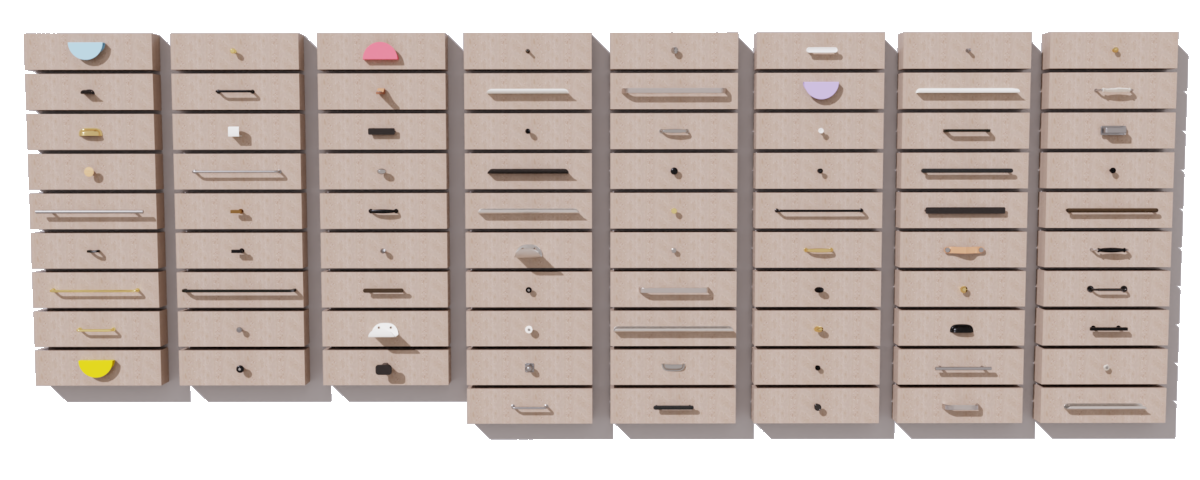}
    \caption{\textbf{Selection of our IKEA Handles Dataset.} We provide 90 geometrically distinct handle assets with multiple texture variations, totaling 190 unique simulation-ready assets. Each handle includes high-resolution collision geometry, realistic textures, and articulated joints with compliant degrees of freedom to support stable grasp synthesis. All assets are based on commercially available products from \url{ikea.com} and can be directly used in Isaac Sim.}
    \label{fig:handles_dataset_overview}
\end{figure*}

\oursubsubsection{Resampling of contact points and grasping commands}
Contrary to analytical optimization approaches where the active contact points $\mathcal{C}=\{c_0, ..., c_{n_c}\}$ are fixed \cite{chen2025bodex} or randomly resampled at each iteration with a fixed probability \cite{li2022gendexgrasp}, we recompute both the assigned contact points and the grasping commands for each new offspring grasp, as shown in Fig.~\ref{fig:hands_overview}.
This adaptive resampling strategy improves efficiency by ensuring that valid grasps are not rejected due to suboptimal contact point selection.

For each grasp proposal $G_i$, we first identify the set of all potential contact points $\mathcal{C}_i$ using a ball query, selecting all points on the object surface within a predefined distance threshold from the hand.
From this candidate set, we select the $n_c$ active contact points $\mathcal{C}_i' \subseteq \mathcal{C}_i$ using farthest point sampling (FPS), which ensures spatially diverse potential contact locations.
Finally, we compute the grasping commands by solving for the joint commands $\Delta\tilde q$ that best achieve the desired normal forces at each contact point:
\begin{equation}
    {\Delta q}_{cmd} = \argmin_{\Delta \tilde{q}} ||J\cdot \Delta \tilde{q} - {N_o}||_2,
\end{equation}
where $J=[J_0, ..., J_{n_c}]$ denotes the stacked contact Jacobians for the active contact points, ${N}_o$ is the desired normal force vector for each contact point, and ${\Delta q}_{cmd}$ represents the commanded change in joint states.





\subsection{Evolutionary Steering and Preference Alignment}
\label{subsec:steering}
An advantage of our evolutionary approach is its flexibility in incorporating diverse steering signals through the fitness function. Unlike gradient-based methods that require differentiable objectives, our evolutionary framework can directly optimize non-differentiable metrics. The fitness function presented in Sec.~\ref{subsec:evolutionary_refinement} already demonstrates this capability: the lifetime metric and the contact distance calculation includes a contact point resampling operation, both of which are not fully differentiable.
In general, we can steer the evolutionary refinement either by (i) modifying tournament selection to incorporate
pairwise preferences, or (ii) augmenting the fitness with absolute, task-specific scores (e.g., grasp energies).


For human preference alignment, we train a PointNet++~\cite{qi2017pointnet} network to predict preference scores for grasp candidates based on hand keypoints and object point cloud configurations, using 1,000 human pairwise annotations. This learned preference model is incorporated directly into the fitness function, biasing the evolutionary process toward grasps that align with human notions of natural or desirable grasping strategies while maintaining physical stability. Importantly, our framework is agnostic to the preference acquisition method -- it can accommodate learned reward models like our PointNet++ predictor, direct human annotation during evolution, or any other preference signal that can score or rank grasp candidates.

\subsection{Diffusion Distillation}
\label{subsec:diffusion_distillation} After generating our grasp pose dataset, we train a diffusion model to enable real-world inference from noisy point clouds.  We build on the DexGraspAnything~\cite{zhong2025dexgrasp} architecture, which diffuses only wrist position and joint states, but extend it to diffuse the complete grasp configuration, including end-effector orientation. We parameterize it via the standard 6-DoF representation using the first two columns of the rotation matrix~\cite{bregier2021deep,geist2024learning}. We further incorporate a differentiable hand model~\cite{graspqp} to compute hand keypoints and surface samples through forward kinematics. This enables geometric supervision during training by defining loss terms not only on noise magnitude but also on keypoint distance and object penetration between ground truth and predicted hand models.

Concretely, as illustrated in the Algorithm ~\ref{alg:grasp_diffusion}, we (i) normalize a ground-truth grasp, (ii) sample a diffusion time step and corrupt the grasp with Gaussian noise, and (iii) train the network to predict the injected noise conditioned on the observed (partial/noisy) point cloud. Beyond the standard denoising objective, we use the hand model to map the predicted grasp back to 3D hand geometry and add two intuitive regularizers: a keypoint consistency loss that keeps the potential hand contact points aligned with those of the target grasp, and a penetration penalty that discourages hand--object interpenetration by penalizing hand surface samples that lie inside the object. Finally, we schedule the collision weight over diffusion steps so that feasibility is emphasized most strongly when the sample is close to the initial sampling distribution, essentially allowing early grasps in the diffusion schedule to be in collision.

\section{Handles Dataset} 
\label{sec:handles}
We provide a total of 90 distinct handle assets collected from the IKEA website (190 when including texture variations), each with detailed collision geometries and high-quality textures suitable for simulation. To ensure stable contact simulation, we re-mesh each collision model by recursively subdividing triangles until a sufficiently high resolution (max 1cm edge length per triangle) is reached. This is important for Isaac Sim's Signed Distance Field (SDF) based collision handling: with coarse collision meshes we observed noticeable interpenetration and unstable contacts. Each handle is attached to a prismatic joint that translates along the z-axis. To introduce mild compliance and encourage force-closure grasps, we additionally add two prismatic joints with small travel limits (\(\pm 1.5\,\mathrm{cm}\), i.e., \(3\,\mathrm{cm}\) total) among the x and y axis. This prevents weak grasps that only ``hold'' the handle by a few millimeters along the compliant direction from being evaluated as stable, and effectively pushes the solutions toward grasps with stronger force-closure properties. We release the final dataset as articulated \texttt{.usd} assets that can be used directly in Isaac Sim. An overview of all handles is shown in Fig.~\ref{fig:handles_dataset_overview}.

\begin{figure*}[ht]
\vspace{-5mm}
\includegraphics[width=1.0\linewidth]{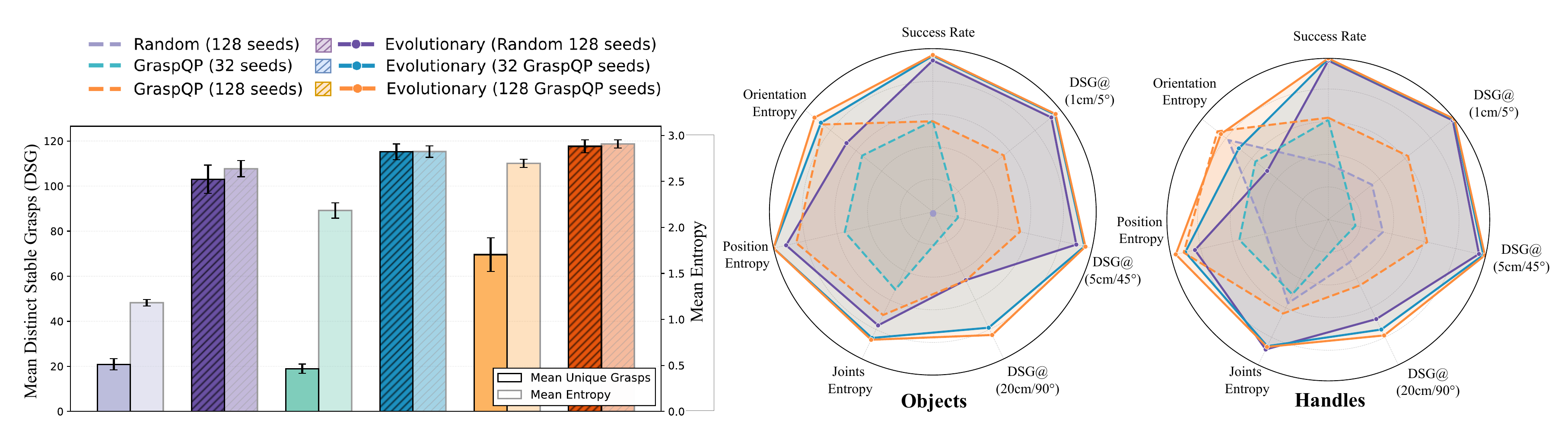}

  \caption{\textbf{Grasp Quality and Diversity Comparison.} Radar plots comparing synthesized grasp distributions across different generation strategies. We report grasp stability (success rate), uniqueness of successful grasps at different thresholds (DSG@2cm/DSG@20cm), and diversity via marginal entropies of hand position, orientation, and joint angles. Overall, evolutionary refinement improves stability while maintaining high diversity across both datasets.}
    \vspace{-5mm}

    \label{fig:combined_spider}
\end{figure*} 
\section{Experiments}
\subsection{Experimental Setup}
\oursubsubsection{Robotic Gripper}
We evaluate our method using the XHand gripper from Robotera. Our pipeline, however, can directly be applied to any gripper supported in IsaacLab.
For the XHand, we use a contact-point set focused on the fingertips, as illustrated in Fig.~\ref{fig:hands_overview}.
To match real-world interaction properties more closely, we define two friction materials per asset: metallic, low-friction surfaces use $\mu = 0.2$, while higher-friction materials (e.g., silicone) use $\mu=0.8$.

\oursubsubsection{Simulation Environment}
For the object evaluation, we use a subset of 50 assets from \cite{wang2023dexgraspnet} and their respective IsaacLab integration from \cite{graspqp}.
For the handles dataset, we selected 50 assets from the handle dataset in Sec.~\ref{sec:handles}, each initialized with the main prismatic joint translating along the
z-axis.

\oursubsubsection{Evolutionary Refinement}
We initialize the evolutionary algorithm with a population size of $S=32$ or $S=128$ and run refinement for a total of $10'000$ evolutionary steps.
We use tournament selection with size $k=4$, a mutation probability of $p_{\text{mutation}}=0.75$, and a crossover probability of $p_{\text{crossover}}=0.2$.
Mutations are applied with Gaussian step sizes $\Sigma=\mathrm{diag}(\Sigma_{\chi},\Sigma_q)$, using $\Sigma_{\chi}=(0.025,\,0.05)$ for the pose components (translation/orientation) and $\Sigma_q=0.04$ for joint states.
For density-aware selection (Eq.~\ref{eq:density-sampling}), we set the neighborhood radius to $r=0.65$ and the power $p=2.0$.
Moreover, because PhysX does not reliably handle assets that are initialized in collision, we perform a few de-penetration steps after mutation to avoid spawning in-collision configurations and ensure valid initialization.
Finally, we cap the archive size to $|\mathcal{P}|\le = 1024$ and, once exceeded, prune by ranking grasps by score and using farthest-point sampling. 

\oursubsubsection{Evaluation Metrics} To assess the performance of our method, we use the following evaluation metrics:

\noindent\underline{Success Rate}: Total ratio of grasps that can successfully hold the object without dropping it during simulation.

 \noindent\underline{DistinctStableGrasps}: The unique grasp metric captures the number of distinct, stable grasps the method can generate. We calculate this metric by discretizing successful grasp poses into bins defined by three resolution parameters: $\delta_r$ for position, $\delta_\phi$ for orientation (Euler angles), and $\delta_q$ for joint states. We then count the number of unique grasps given this quantization and report DistinctStableGrasps (DSG) as the total number of unique, stable grasps at the specified resolution. We evaluate at three resolutions: DSG@(20cm, 90°, 90°), DSG@(2cm, 45°, 45°) and DSG@(1cm, 5°, 5°).

\noindent \underline{Entropy (H)}: For each object, we calculate the joint entropy of the distribution of successful grasps over position, orientation, and joint states as done in  \cite{graspqp}. We report either the average entropy over position, orientation, and joint states, or each value independently, depending on the analysis context.

\oursubsubsection{Evaluation Procedure} We assess the stability of each grasp proposal using Isaac Lab~\cite{mittal2025isaac} following prior evaluation protocols~\cite{wang2023dexgraspnet,graspqp}, i.e., for each asset we reset the simulation, place the hand at the proposed wrist pose, execute the corresponding grasp closing command, and test stability under external disturbances over a fixed rollout horizon.
For the \emph{Objects} dataset, we follow the standard disturbance test by sequentially applying forces along the canonical axes and declaring success if the object remains stably grasped (within the allowed pose thresholds) throughout at least one of the three axes.
For the \emph{Handles} dataset, we evaluate stability by applying a pulling force along the compliant direction (negative $z$-axis) and deem a grasp successful if the handle remains stably held throughout the rollout.

\subsection{Influence of Evolutionary Refinement}

To systematically assess the impact of evolutionary refinement on grasp quality and diversity, we analyze grasp distributions generated by different initialization strategies on representative test sets from both the \emph{Objects} and \emph{Handles} datasets (Fig.~\ref{fig:combined_spider}). Our investigation addresses two key questions: (1) whether evolutionary refinement improves grasp stability while maintaining diversity, and (2) how the choice of initial population affects the final grasp distribution.


Our results reveal several important findings. First, evolutionary refinement consistently and substantially improves both stability and unique grasp coverage across all initialization strategies and both datasets, elevating success rates and DSG@(<20cm,<90°) metrics close to saturation. Critically, these gains in stability do not come at the expense of diversity. The refined distributions preserve, and often increase, the entropy across pose and joint spaces. Second, while increasing the number of analytically generated seeds (using GraspQP) from 32 to 128 improves the baseline, evolutionary refinement consistently outperforms unrefined methods at both seed counts. This demonstrates that simulator-in-the-loop optimization is essential for generating physically feasible grasps beyond what analytical methods can achieve alone.

Perhaps most surprisingly, evolutionary refinement starting from random initialization achieves competitive success rates and unique grasp counts, even when the initial random distribution yields zero successful grasps (as observed in the Objects dataset). This robustness underscores our method's capacity to synthesize diverse, stable grasps from highly suboptimal starting points. However, this comes with a notable trade-off: randomly initialized refinement exhibits markedly lower entropy, particularly in orientation and, to a lesser extent, joint configurations. This suggests that while our evolutionary approach can recover stability from poor initializations, analytical seeding remains crucial for maintaining a rich, multimodal grasp distribution that spans the full space of feasible solutions.

\begin{figure}[t!]
    \centering
    \includegraphics[width=1.0\linewidth]{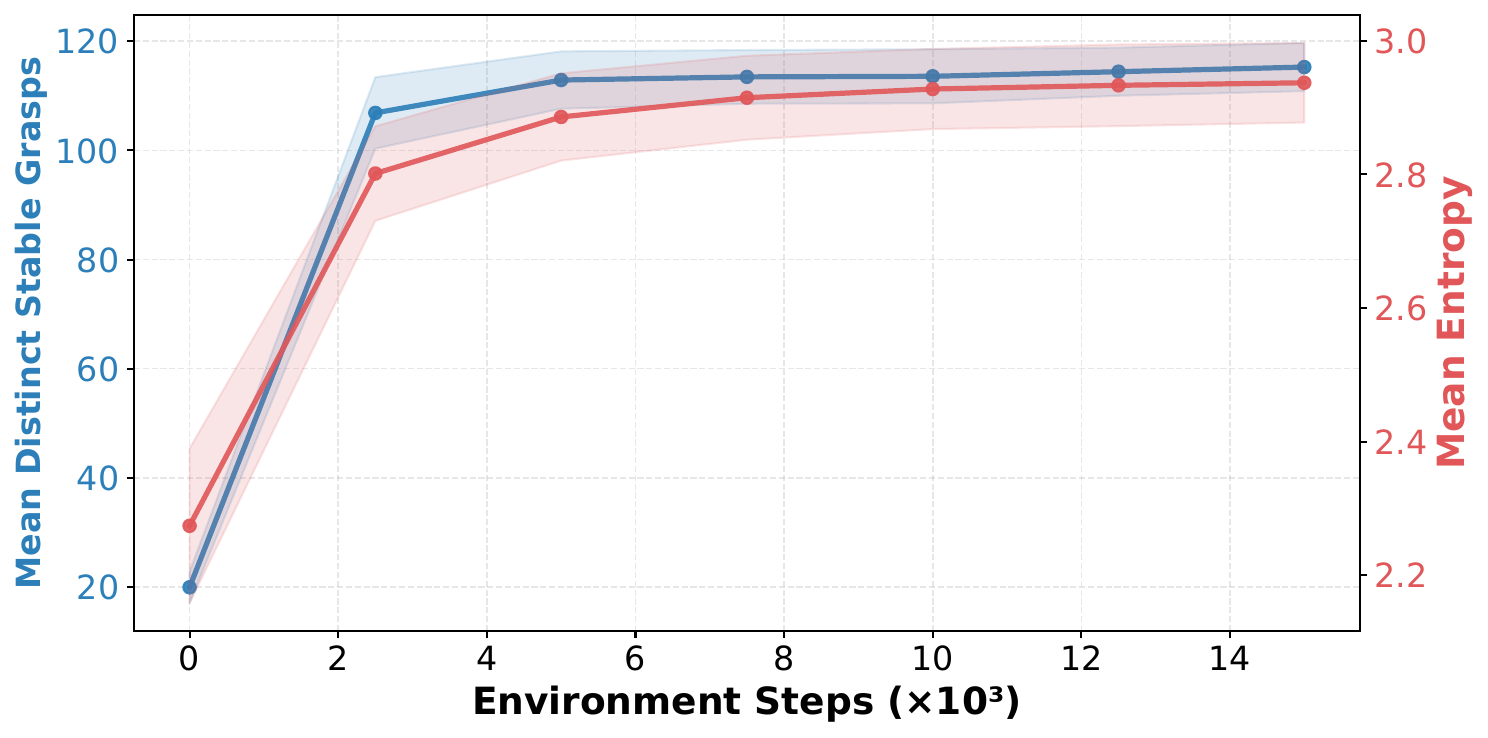}
     \caption{\textbf{Convergence of Evolutionary Refinement.} Mean distinct stable grasps (blue, left axis) and mean entropy (red, right axis) as a function of simulator environment steps, initialized with 32 analytical seeds. Both metrics exhibit rapid improvement during the first 2.5k steps, then converge to a stable plateau by approximately 10k steps. The simultaneous saturation of both stability and diversity metrics shows that evolutionary refinement efficiently balances grasp quality and coverage within a limited interaction budget.}
     \label{fig:convergence}
     \vspace{-3mm}
\end{figure}

\oursubsubsection{Convergence Behavior}
To characterize the sample efficiency and convergence properties of our evolutionary refinement approach, we visualize the grasp quality and diversity metrics as a function of simulator environment steps in Fig.~\ref{fig:convergence}. We initialize the population with 32 analytical seeds and monitor both the mean number of unique successful grasps (blue, left axis) and mean entropy across position, orientation, and joint configuration spaces (red, right axis) throughout the optimization process.

This convergence behavior demonstrates two key properties of our approach. First, evolutionary refinement is highly sample-efficient, achieving near-optimal performance within a limited interaction budget of 10k steps. Second, the method reliably converges to distributions that simultaneously maximize grasp stability and preserve diversity, avoiding common failure modes such as premature convergence to a single grasp mode or uncontrolled diversity at the expense of physical feasibility.

\subsection{Effect of Training a Diffusion Model}

A natural alternative to simulator-in-the-loop evolutionary refinement is to leverage generative modeling: train a diffusion model on the analytically-seeded grasp distribution and sample novel grasps from the learned implicit manifold. To evaluate this approach, we train a conditional diffusion model on the initial seed distribution (both 32 and 128 seeds) from the \emph{Objects} dataset and compare the quality and diversity of generated grasps against both the raw analytical seeds and evolutionary refinement.

\begin{figure}[t!]
    \centering
    \includegraphics[width=1.0\linewidth]{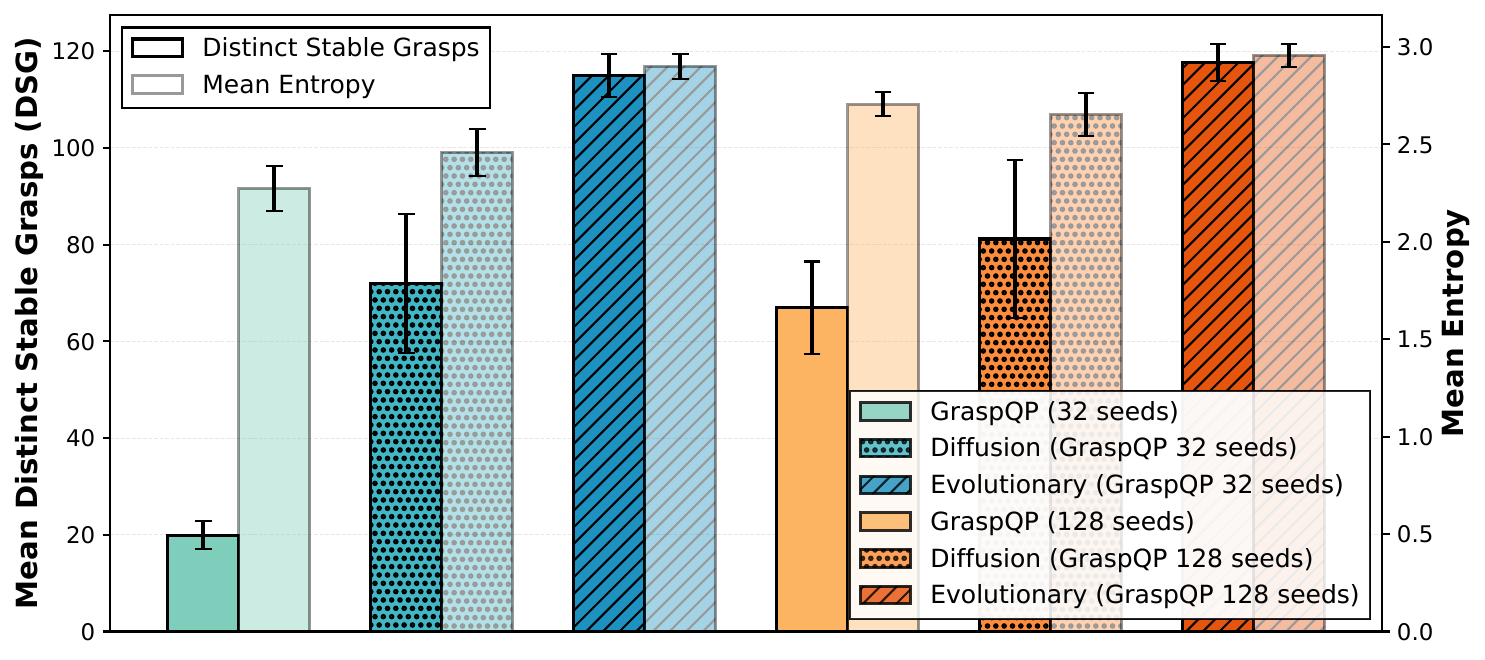}
    \caption{\textbf{Effect of Diffusion Training vs Evolutionary Refinement.} Mean distinct stable grasps for GraspQP, a diffusion model trained on GraspQP samples, and evolutionary refinement, showing that diffusion improves over the raw initialization but remains below evolutionary sampling.}
\vspace{-5mm}
    \label{fig:diffusion_training}
\end{figure}

Fig.~\ref{fig:diffusion_training} presents mean unique successful grasps (solid bars, left) and mean entropy (hatched bars, right) for analytical, diffusion, and evolutionary grasp synthesis. The results reveal several key findings. First, diffusion improves upon the raw analytic initialization with GraspQP, increasing mean unique grasps from approximately 20 to 71 (32 seeds) and from 70 to 82 (128 seeds), while maintaining or slightly increasing entropy. This demonstrates that the diffusion model successfully learns to interpolate within the analytical grasp manifold, generating configurations that expand coverage of the feasible grasp space.

However, diffusion consistently underperforms evolutionary refinement across all settings. With 32 seeds, evolutionary refinement achieves approximately 115 unique grasps compared to diffusion's 72 -- a 60\% improvement. With 128 seeds, the gap narrows but remains substantial (118 vs. 81 unique grasps, a 46\% improvement). Critically, evolutionary refinement also achieves a higher average entropy (approximately 3.0) compared to diffusion (approximately 2.4–2.6), indicating that the performance gap manifests in both grasp stability and diversity. As evolutionary refinement increases the entropy of the initialization set, this suggests it can extrapolate and expand beyond the original distribution. This aligns with evolutionary refinement's ability to maintain exploration across diverse modes while enforcing physical constraints, resulting in both higher success rates and greater distributional spread. Diffusion is most effective in the low-sampling regime (32 seeds), where it provides substantial gains over raw initialization despite limited training data. However, for achieving maximum grasp quality and uniqueness, direct optimization in the high-fidelity simulator through evolutionary refinement remains the superior approach, as it explicitly enforces physical feasibility through contact dynamics rather than relying on implicit learned priors. Nonetheless, it remains essential for efficient deployment from partial observations.

\subsection{Preference-Guided Evolutionary Refinement}
While evolutionary refinement generates diverse, stable grasps, the resulting configurations may not align with human preferences or task-specific requirements for natural grasping strategies. To address this, we incorporate human feedback by training a reward model on preference comparisons between grasp pairs and using it to bias the fitness evaluation during evolutionary refinement (see Section~\ref{subsec:steering}).

\begin{figure}[t!]
    \centering
    \vspace{-5mm}
    \includegraphics[width=0.9\linewidth]{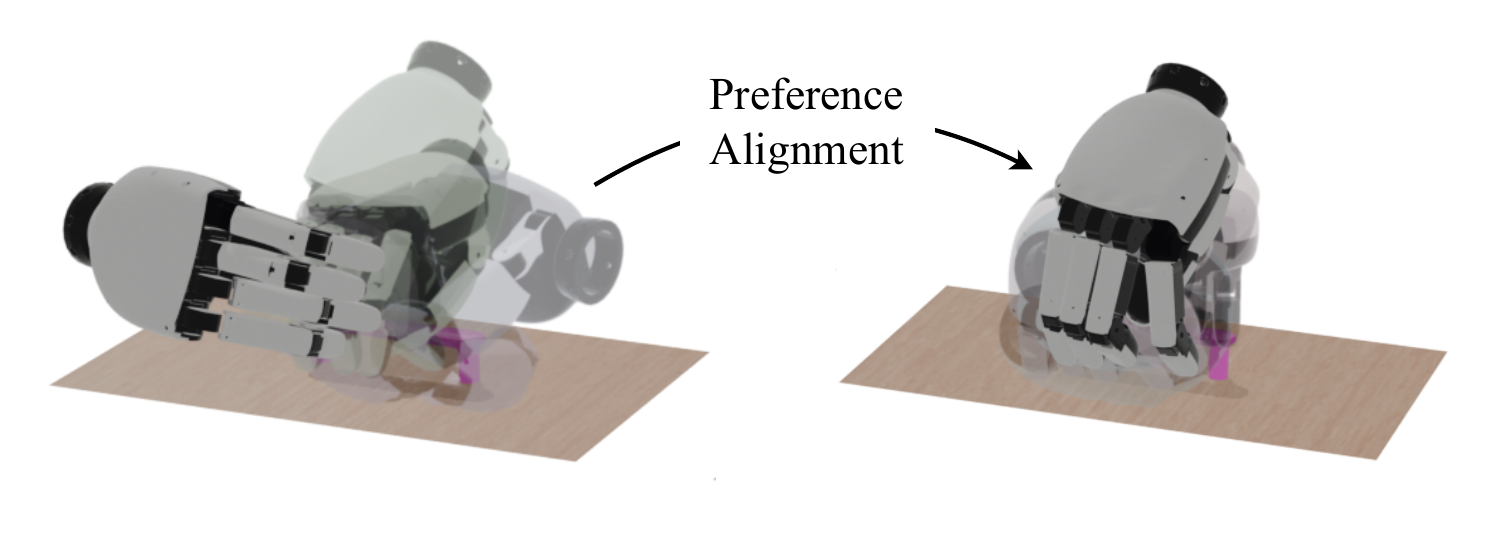}
    \caption{\textbf{Effect of Preference Alignment.} Comparison of grasps without (left) and with (right) preference alignment from human feedback. The un-guided evolutionary distribution produces diverse but unconventional grasps, while the preference-aligned version generates grasps that better match human intent and natural grasping patterns.}
    \label{fig:grasp_visuals}
    \vspace{-5mm}
\end{figure}

We collect approximately 1,000 human pairwise preference annotations on grasp candidates, where annotators compare two grasps and select which appears more natural and socially appropriate. Preferences are based on intuitive human grasping behavior -- for instance, discouraging grasps that might be functional but unnatural, such as grasping a handle between the pinky and thumb or approaching from awkward angles. We train a PointNet++ network~\cite{qi2017pointnet} that takes as input the object point cloud and hand keypoint configuration and outputs a preference score. The network is trained with a standard Bradley-Terry pairwise ranking loss to predict the probability that one grasp is preferred over another. During evolutionary refinement, we add the predicted preference score to the fitness function.
Fig.~\ref{fig:grasp_visuals} illustrates the effect of preference alignment. Without guidance (left), evolutionary refinement produces physically stable but unconventional grasps with awkward approach angles and unintuitive finger placements. With preference alignment (right), the method generates more natural configurations that align with typical human grasping patterns, approaching from conventional orientations with expected finger placements while maintaining physical stability. This demonstrates that preference guidance can effectively shape grasp distributions toward human-like behavior without sacrificing feasibility.

\subsection{Real-World Robotic Deployment}

\begin{figure}
    \centering
    \vspace{-5mm}
   \includegraphics[width=0.9\linewidth]{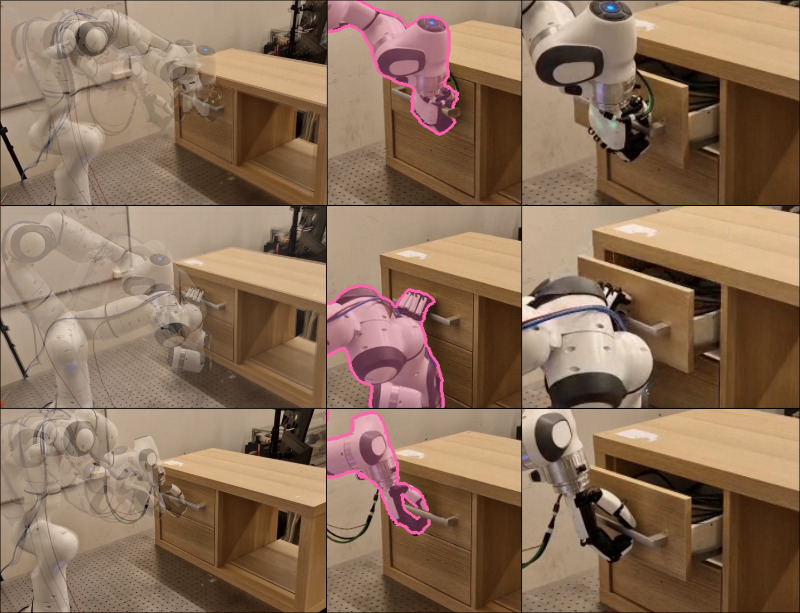}
    \caption{\textbf{Real-world Cabinet-Handle Rollout.} We sample handle grasps from a point-cloud-conditioned diffusion model trained on refined grasp distributions, then plan collision-aware trajectories using cuRobo. These are executed on a Franka Panda arm equipped with a dexterous hand. Three executions across different grasp modes and approach directions demonstrate successful sim-to-real transfer in contact-rich manipulation.}
    \label{fig:rollout_example}
    \vspace{-5mm}
\end{figure}

To validate the real-world transferability of our approach, we deploy synthesized grasps on a physical robotic system consisting of a Franka Panda arm equipped with a dexterous XHand. For a given target cabinet, we first capture ten RGB-D frames using an Intel RealSense L415 camera. We use Depth Anything V3~\cite{depthanything3} to lift the RGB images and obtain depth predictions, which we co-align with the sensor depth measurements to ensure consistent scale and registration. This alignment process produces a coherent 3D point cloud reconstruction of the scene, even when objects exhibit specular or transparent surfaces that challenge traditional depth sensors. We then downsample the point cloud by voxelizing it and use it to sample grasps from a diffusion model trained on our refined handle grasp distributions, with collision-aware gradient guidance enabled to bias sampling toward collision-free configurations. Finally, we use cuRobo~\cite{sundaralingam2023curobo} to generate kinematically feasible manipulation trajectories, approximating the robot as a set of spheres and the environment as the voxelized point cloud. These trajectories are executed using a joint impedance controller on the physical hardware.
Fig.~\ref{fig:rollout_example} shows three successful execution sequences on a cabinet with handles, demonstrating diverse approach strategies across different grasp modes. The successful executions across multiple grasp modes demonstrate that our method transfers effectively from simulation to real hardware. Additional hardware experiments and extended rollout sequences are provided in the supplementary video and Appendix.

\section{Limitations}
\label{sec:limitations}
While our approach demonstrates significant improvements in grasp stability and diversity, several limitations remain. First, evolutionary refinement still benefits from being initialized with a somewhat reasonable grasp distribution: while our method can recover from suboptimal initializations (e.g., random seeds), the refined distribution exhibits reduced diversity compared to analytically seeded runs, suggesting that strong initialization improves the final coverage of the grasp manifold. 
Second, PhysX does not reliably resolve contacts when assets are initialized in collision; as a result, we perform explicit de-penetration steps after mutation to ensure valid initialization, which adds non-trivial computational overhead.
 Third, our real-world deployment pipeline is relatively slow due to its multi-stage nature (multi-view RGB-D capture, depth lifting/alignment, point-cloud processing, diffusion sampling, and motion planning), which limits end-to-end responsiveness.


\section{Conclusion}
\label{sec:conclusion}
We presented \textsc{DexEvolve}, a scalable generate-and-refine pipeline for synthesizing large-scale, diverse, and physically feasible dexterous grasps. Starting from analytically generated seeds, we refine grasp configurations directly in a high-fidelity, non-differentiable simulator using an asynchronous evolutionary algorithm with archive-based novelty insertion and density-aware selection, enabling efficient exploration while suppressing mode collapse. Across both the \emph{Objects} and \emph{Handles} datasets, evolutionary refinement substantially improves grasp stability and unique grasp coverage, achieving over 120 distinct stable grasps per object, a 1.7 - 6x improvement over the analytical seeds -- while maintaining high entropy across position, orientation, and joint configuration spaces. The refined distributions consistently outperform both unrefined analytical methods and diffusion-based alternatives in terms of physical feasibility and distributional diversity.

Beyond synthesis performance, our framework supports gradient-free steering through non-differentiable objectives, enabling preference alignment with human feedback and task-specific quality metrics. We validate real-world transferability through successful deployment on physical hardware and distill the refined grasp distribution into a point-cloud-conditioned diffusion model to enable fast inference for real-world deployment. 

. 

\bibliographystyle{plainnat}
\bibliography{references}

\newpage
\begin{figure*}[t!]
    \centering
    \includegraphics[width=1.0\linewidth]{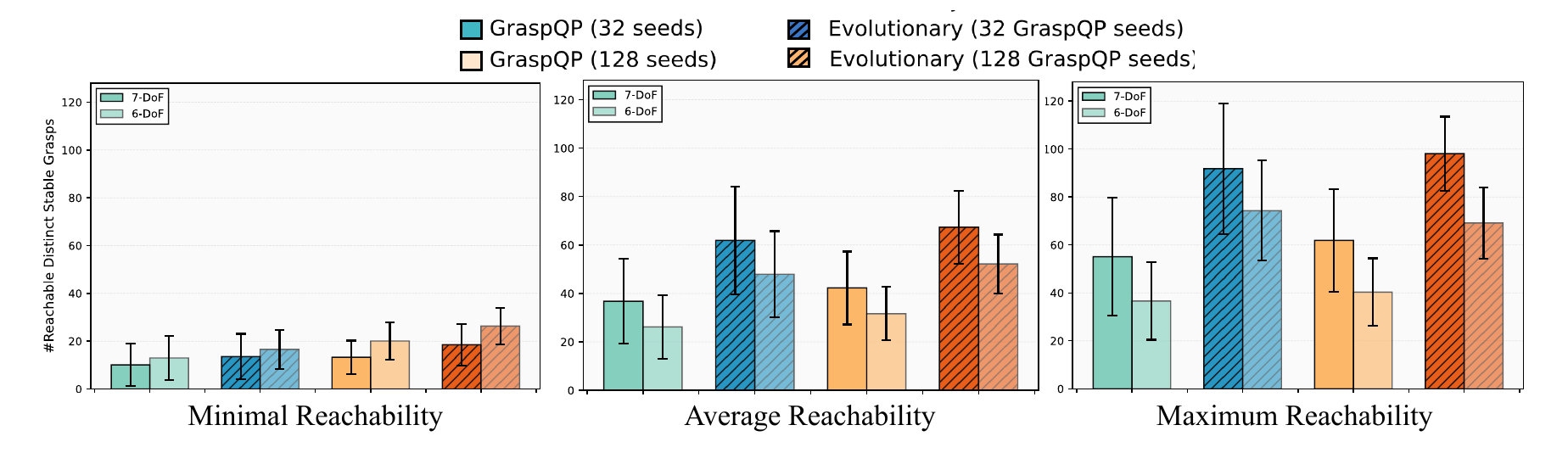}%
    \caption{\textbf{Reachability analysis comparing 7-DoF and 6-DoF robot arms.} We compare the number of distinct stable grasps achievable across different handle positions. Grasp candidates are generated by GraspQP (solid colors) and our evolutionary approach (hatched) using 32 and 128 seed grasps on 50 handle assets. For 12 sampled handle positions, we report: (a) \textbf{Minimal Reachability:} mean minimum reachable grasps across all object poses; (b) \textbf{Average Reachability:} mean reachable grasps across all poses; (c) \textbf{Maximum Reachability:} mean maximum reachable grasps for the most favorable pose. Error bars indicate standard deviation.}
    \label{fig:reachability_analysis}
\end{figure*}

\section{Supplementary Material}

\subsection{Evolutionary Refinement: Additional Algorithmic Details}
\label{sec:evolutionary_details}

\parag{State and parameters.}
We represent each individual as $G = (\chi, q, \Delta q_{cmd})$ (Sec.~\ref{subsec:evolutionary_refinement}), where $\chi\in SE(3)$ is the wrist pose, $q\in\mathbb{R}^{12}$ are the hand joint angles, and $\Delta q_{cmd}\in\mathbb{R}^{12}$ is the closing velocity command.
We cap the archive size at $P_{\max}=1024$ and run refinement for $T=10{,}000$ environment steps. Unless otherwise noted, the simulator rollout uses a maximum lifetime horizon of 3 evaluation windows (as described in Section~\ref{subsec:evolutionary_refinement}). When considering mutation and/or insertion into the archive, we ignore the closing velocities as they will be re-computed anyway.

\parag{Distance metric and novelty threshold.}
For novelty gating and duplicate suppression, we embed each grasp as $\phi(G) = [\, 10\,p,\; r,\; q\,]$, where $p\in\mathbb{R}^3$ is wrist position, $r\in\mathbb{R}^3$ are Euler angles (converted from the wrist quaternion), and $q$ are joint positions.
This corresponds to scaling translation by 10 (to balance its contribution with orientation and joints in the distance metric) and leaving orientation/joints unscaled.
Given two grasps $G,G'$, we compute
\begin{equation}
\label{eq:appendix_distance}
\begin{aligned}
    d(G,G')
    &= \tfrac{1}{2}\sqrt{\mathrm{mean}\!\left((\phi(G)_{1:7}-\phi(G')_{1:7})^2\right)} \\
    &\quad + \tfrac{1}{2}\sqrt{\mathrm{mean}\!\left((\phi(G)_{8:}-\phi(G')_{8:})^2\right)}.
\end{aligned}
\end{equation}
A candidate is considered too close to the archive if $\min_{G'\in\mathcal{P}} d(G,G') < \tau$ with $\tau=0.1$. When the archive size exceeds $P_{\max}=1024$, we prune to 75\% of maximum capacity (768 grasps) by keeping the highest-scoring individuals.

\parag{Selection and density parameters.}
We use tournament selection with tournament size $k_{\text{tourn}}=4$.
For density-aware selection, we use radius $r=0.65$ and exponent $p=2.0$ for the density suppression term.

\parag{Mutation and crossover.}
We use mutation probability $p_{\text{mutation}}=0.75$ and crossover probability $p_{\text{crossover}}=0.2$.
Mutation is applied in Euler pose space with Gaussian step sizes
$\Sigma_{\chi}=(0.025,0.05)$ (translation in meters/orientation in radians) and $\Sigma_q=0.04$ for normalized joint angles.
We use hand-pose crossover, swapping at the fixed split between pose and joints. To mitigate in-collision initialization failures, we apply up to $N_{\text{depenetration}}=2$ explicit de-penetration steps after mutation. Each step moves the hand along the negative gradient of the penetration energy by step size $\alpha_{\text{depenetration}}=0.15$. To prevent large steps, gradients are clipped: pose translation to $[-0.1, 0.1]$, orientation to $[-0.05, 0.05]$, and joints to $[-0.5, 0.5]$.
\begin{algorithm}[t!]
\footnotesize
\setlength{\interspacetitleruled}{2pt}
\setlength{\interspacealgoruled}{2pt}
\SetAlgoLined
\KwIn{Seed population $\mathcal{P}$ (size $S$), archive cap $P_{\max}$, steps $T$, number of returned grasps $k=128$}
\KwOut{Refined archive $\mathcal{P}$ and final successful set $G^*$}

\BlankLine
$(\mathcal{P}_A, \mathcal{F}_A) \leftarrow \textsc{InitializeEnvs}(\mathcal{P})$\;
\For{$t \leftarrow 1$ \KwTo $T$}{
    \textsc{AdvanceSim}(\ensuremath{\mathcal{P}_A})\;
    $(\mathcal{P}', (E_{\text{lifetime}}', E_{\text{dis}}', E_{\text{pen}}')) \leftarrow \textsc{CollectFinished}(\ensuremath{\mathcal{P}_A})$\;
    $\mathcal{F}' \leftarrow E_{\text{lifetime}}' - w_{\text{dis}} E_{\text{dis}}' - w_{\text{pen}} E_{\text{pen}}'$\;

    \tcp{Archive update: novelty gate + local competition (Sec.~\ref{subsec:evolutionary_refinement})}
    \ForEach{$(G, \mathcal{F}(G)) \in (\mathcal{P}', \mathcal{F}')$}{
        $\mathcal{P} \leftarrow \textsc{InsertOrReplace}(\mathcal{P}, G, \mathcal{F}(G))$\;
    }
    \If{$|\mathcal{P}| > P_{\max}$}{
        $\mathcal{P} \leftarrow \textsc{PruneTopScore}(\mathcal{P}, \lfloor 0.75\,P_{\max}\rfloor)$ \tcp*{keep best 75\% by score}
    }

    \tcp{Generate new batch of candidates}
    \For{$i \leftarrow 1$ \KwTo $|\mathcal{P}'|$}{
        $F_d \leftarrow \textsc{DensitySuppress}(\mathcal{F}_i, \mathcal{F}_A')$\;
        $(G_1, G_2) \leftarrow \textsc{TournamentSelect}(\mathcal{P}, \mathcal{F}_d)$\;
        $\tilde{G} \leftarrow \textsc{Crossover}(G_1, G_2)$\;
        $\tilde{G} \leftarrow \textsc{Mutate}(\tilde{G})$\;
        $\tilde{G} \leftarrow \textsc{Depenetrate}(\tilde{G})$\;

        \tcp{Resample contacts + closing command}
        $\mathcal{C} \leftarrow \textsc{BallQueryContacts}(\tilde{G})$\;
        $\mathcal{C}' \leftarrow \textsc{FPS}(\mathcal{C}, n_c)$\;
        ${\Delta q}_{cmd} \leftarrow \argmin_{\Delta \tilde{q}} \lVert J(\mathcal{C}')\,\Delta \tilde{q} - N_o(\mathcal{C}') \rVert_2$\;
        $\tilde{G} \leftarrow (\tilde{\chi}, \tilde{q}, {\Delta q}_{cmd})$\;

        $\mathcal{P}_A[i] \leftarrow \tilde{G}$\;
    }
}
\tcp{Return a final grasp set from the archive}
$\mathcal{G}_{\text{succ}} \leftarrow \{G \in \mathcal{P} : E_{\text{lifetime}}(G) \ge E_{\min}\}$\;
$G^* \leftarrow \textsc{SelectTopOrFPS}(\mathcal{G}_{succ}, k)$\;
\Return{$(\mathcal{P}, G^*)$}\;
\caption{Asynchronous evolutionary refinement (high-level pseudo-code).}
\label{alg:evolutionary_algorithm}
\end{algorithm}

\parag{Energy weights.}
The default energy weights are set to $w_{\text{dis}}=1.0$, $w_{\text{pen}}=100.0$, and $w_{\text{lifetime}}=1.0$. When preference guidance is enabled, we set $w_{\text{reward}}=10.0$; otherwise $w_{\text{reward}}=0$.

\parag{Penetration term $E_{\text{pen}}$.}
For each environment we evaluate penetration between the hand and the object using two complementary signals. For a candidate grasp we compute: (i) penetration depths $\mathbf{p}\in\mathbb{R}^{N_o}$ from hand geometry against object surface points, and (ii) negated SDF values $\mathbf{s}\in\mathbb{R}^{N_s}$ from object geometry evaluated at hand surface points. 
We then form
\begin{equation}
\label{eq:appendix_penetration}
    E_{\text{pen}}
    = \sum_{j} \max\bigl(0,\; \delta - p_j\bigr) + \sum_{i} \max\bigl(0,\; \delta - s_i\bigr),
\end{equation}
where $\delta=3\,\mathrm{mm}$ is a small collision offset used to account for minor penetrations at contact and differences between the analytical SDF model and the Isaac Sim model.

In the fitness, this enters as a penalty term $-w_{\text{pen}}\,E_{\text{pen}}$.

\parag{Distance term $E_{\text{dis}}$.}
We compute $E_{\text{dis}}$ as an SDF-based distance-to-contact signal: we evaluate the object SDF at the hand contact points and use $|\mathrm{SDF}|$ as the unsigned distance to the object surface.
We then downsample the contact candidates to a fixed number of contact points $n_c=12$ (thresholding + farthest point sampling) and aggregate the resulting distances into a scalar that penalizes grasps far from contact-feasible configurations.
When enabled, this enters the fitness as $-w_{\text{dis}}\,E_{\text{dis}}$.

\parag{Contact point computation.}
For each candidate grasp, we identify potential contact points by performing a ball query with radius $r_{\text{ball}}=0.025\,\mathrm{m}$ around hand surface points to find contact points that are close enough to the object.
After identifying contact candidates, we apply farthest point sampling (FPS) to select the most diverse $n_c=12$ contacts for computing the closing command $\Delta q_{cmd}$.
The Jacobian $J(\mathcal{C}')$ is computed analytically from the hand's kinematic model, and contact normals $N_o(\mathcal{C}')$ are obtained from the object SDF gradients.

\subsection{Pseudo-code: Asynchronous Evolutionary Loop}
\label{sec:pseudocode}
Algorithm~\ref{alg:evolutionary_algorithm} summarizes our asynchronous simulator-in-the-loop evolutionary refinement. We maintain an archive $\mathcal{P}$ of grasp candidates and iteratively (i) select parents using density-aware tournament selection, (ii) generate offspring via structured crossover and mutation (with de-penetration), (iii) evaluate offspring in parallel Isaac Sim rollouts to obtain $E_{\text{lifetime}}$, $E_{\text{dis}}$, and $E_{\text{pen}}$, and (iv) update the archive using novelty gating with threshold $\tau=0.1$.

\begin{figure*}
    \centering
    \includegraphics[width=1.0\linewidth]{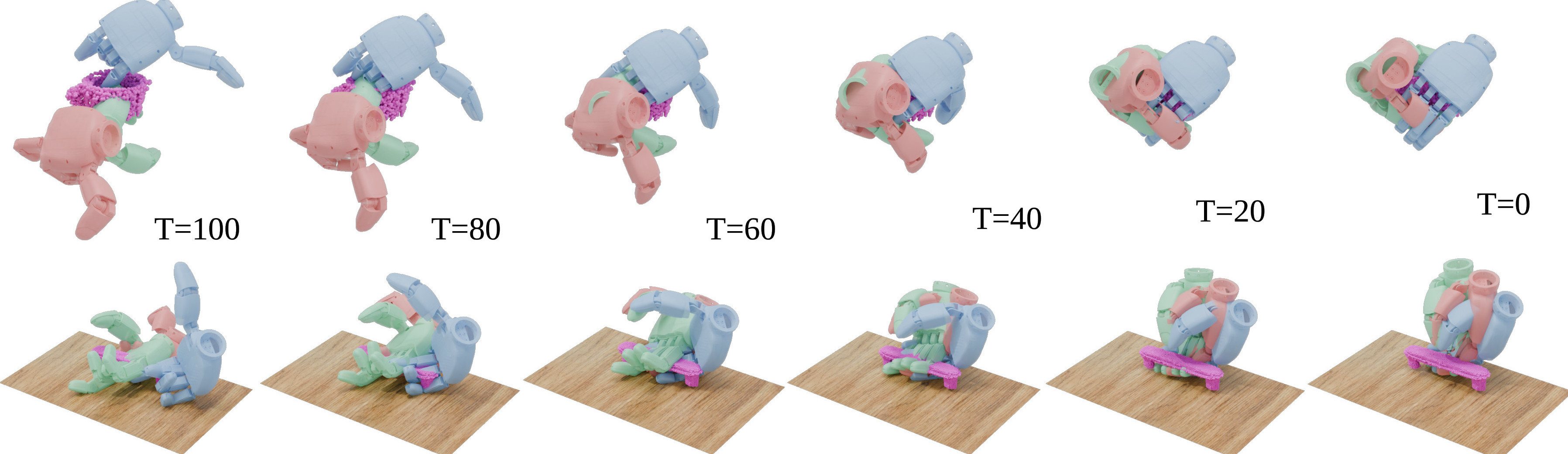}
   \caption{\textbf{Diffusion-based grasp generation.} Visualization of the denoising process for grasp pose prediction on a cup (top row) and bar-style handle (bottom row). The sequence shows the evolution from the initial noisy distribution (T=100) to the final predicted grasp pose (T=0) at 20-step intervals for three different grasp samples.}

    \label{fig:diffusion_denoising}
\end{figure*}

\subsection{Diffusion Model: Architecture and Training}
\label{sec:diffusion_model}
Figure~\ref{fig:diffusion_denoising} visualizes the diffusion denoising process for grasp pose prediction on various objects, showing the progressive refinement from noisy initialization to the final grasp pose.

\parag{Overview.}
Our diffusion network builds upon the DexGraspAnything architecture~\cite{zhong2025dexgrasp}, extending it in three key ways to improve grasp quality and sim-to-real transfer: (i) we diffuse the full wrist pose (position \emph{and} orientation), not only position; (ii) we condition on both the observed object point cloud and hand keypoints (obtained via forward kinematics); and (iii) we add geometry-based guidance losses (keypoint consistency and penetration penalties) to improve contact consistency and sim-to-real robustness.

\parag{Diffused state.}
We diffuse a $21$-dimensional grasp vector $\mathbf{x}\in\mathbb{R}^{21}$ consisting of wrist translation, a 6D rotation representation, and joint angles,
\begin{equation}
    \mathbf{x} = [x,y,z,\; r_{11},r_{21},r_{31},\; r_{12},r_{22},r_{32},\; q_1,\ldots,q_{12}].
\end{equation}
Here, $(x,y,z)$ is the wrist translation, $(r_{11},r_{21},r_{31})$ and $(r_{12},r_{22},r_{32})$ denote the first two columns of the wrist rotation matrix (a continuous 6D orientation representation), and $q_1,\ldots,q_{12}$ are the hand joint angles.

\parag{Normalization strategy.}
To improve training stability and generalization, we normalize the diffused grasp parameters as follows:
\begin{itemize}
    \item \textbf{Wrist position:} Normalized by a fixed scale of 0.1m (approximately the workspace radius).
    \item \textbf{Wrist orientation:} We normalize the first two columns of the orientation matrix.
    \item \textbf{Hand joints:} Normalized per-joint by their physical limits $[q_{\min,i}, q_{\max,i}]$ to the range $[-1,1]$ via $\tilde{q}_i = 2(q_i - q_{\min,i})/(q_{\max,i} - q_{\min,i}) - 1$.
\end{itemize}

\parag{Conditioning input.}
The conditioning signal consists of an observed point cloud $\mathcal{P}_{obs}$ (voxel-downsampled with normals) concatenated with all potential hand keypoints $\mathcal{K}(\mathbf{x})$ computed from the current (noisy) grasp state via forward kinematics. In our experiments we use a total of 2048 points after concatenation and add an additional binary channel to indicate if a point comes from the point cloud ($=0$) or from a keypoint ($=1$).
The object point cloud is downsampled to 1976 points and we use 72 hand keypoints on the full hand.

\parag{Architecture.}
We use the same architecture as \cite{zhong2025dexgrasp} but reduce the parameter count by downscaling the layer sizes. Our architecture consists of (i) a PointTransformerV3 encoder that processes the conditioning point cloud to extract geometric features, and (ii) a Diffusion Transformer (DiT) that predicts the noise $\hat{\boldsymbol{\epsilon}}_\theta$ given the noisy grasp state, timestep, and encoded features.
The total parameter count is approximately 4.9M parameters for PointTransformerV3 and 4.8M parameters for the DiT, totaling 9.7M parameters.

\parag{Augmentations for sim-to-real.}
To improve robustness to sensor noise and partial observability, we apply several point-cloud augmentations during training:
\begin{itemize}
\item \textbf{Point position noise:} To simulate depth sensor noise, we apply additive Gaussian noise $\text{noise}\sim\mathcal{N}(0, \sigma_{\text{pos}}^2)$ with $\sigma_{\text{pos}}=2\,\mathrm{mm}$. We bias the noise toward the point normals, using $p_{\text{noise}} = p_{\text{gt}} + (0.9 \cdot \vec{n}_p + 0.1) \cdot \text{noise}$ and truncate it with max limit of $\pm 5\,\mathrm{mm}$.
\item \textbf{Normal noise:} To account for normal estimation errors, we perturb point normals by adding noise to the normal direction with standard deviation $\sigma_{\text{n}}=0.05$ before re-normalizing.
\item \textbf{Occlusion:} To simulate partial observability, we randomly select three spheres with radius $2.5\,\mathrm{cm}$ and remove all points that lie within those spheres.
\item \textbf{Global SE(3) augmentation:} To ensure pose-invariant predictions, we apply random rotations uniformly sampled from $SO(3)$ and translations sampled from $\mathcal{N}(0, \sigma_{\text{trans}}^2 I_3)$ with $\sigma_{\text{trans}}=10\,\mathrm{cm}$ for object grasps. For handles, we restrict rotations around the handle axis (z-axis) with angles uniformly sampled from $[-180^\circ, +180^\circ]$.
\end{itemize}

\parag{Training objective.}
Our training objective combines three complementary terms (see Algorithm~\ref{alg:grasp_diffusion}): (i) a standard diffusion denoising loss ($\mathcal{L}_\epsilon$), (ii) keypoint consistency supervision ($\mathcal{L}_k$) comparing predicted and ground-truth hand keypoints via forward kinematics, and (iii) hand--object interpenetration penalties ($\mathcal{L}_p$) using SDF-based penetration signals. The penetration loss is scheduled with an exponential decay that applies stronger penalties at later (less noisy) diffusion steps. The final loss is $\mathcal{L}_{\text{total}} = \lambda_\epsilon \mathcal{L}_\epsilon + \lambda_k \mathcal{L}_k + \lambda_p \mathcal{L}_p \cdot e^{-\frac{t}{T_{\text{max}}/2}}$ with weights $\lambda_\epsilon = 1.0$, $\lambda_k = 1.0$, and $\lambda_p = 0.25$.

\parag{Closing velocity / command computation.}
At inference time, after sampling a grasp pose $\mathbf{x}$, we compute the closing command by querying candidate hand contact points (ball query), selecting active contacts (FPS), and solving the same least-squares command used in the evolutionary stage (Sec.~\ref{subsec:evolutionary_refinement}).

\begin{algorithm}[b!]
\footnotesize
\setlength{\interspacetitleruled}{2pt}
\setlength{\interspacealgoruled}{2pt}
\caption{Diffusion-based Training for Robotic Grasp Pose Generation}
\label{alg:grasp_diffusion}
\KwIn{Ground truth grasp pose $\mathbf{x}_{\text{gt}} \in \mathbb{R}^d$ (position, orientation, joint angles)}
\KwIn{Ground truth object point cloud $\mathcal{P}_{\text{gt}} = \{\mathbf{p}_i\}_{i=1}^N$}
\KwIn{Noisy observations $\mathcal{P}_{\text{obs}}$ with normals $\{\mathbf{n}_i\}_{i=1}^N$}
\KwIn{Differentiable hand model $\mathcal{H}: \mathbb{R}^d \rightarrow \mathbb{R}^{3 \times K}$}
\KwIn{Number of diffusion steps $T$, loss weights $\{\lambda_p, \lambda_k\}$}
\KwOut{Composite training loss $\mathcal{L}_{\text{total}}$}
\BlankLine

\tcp{Data normalization}
$\mathbf{x}_{\text{gt}}^{\text{norm}} \gets \mathrm{Normalize}(\mathbf{x}_{\text{gt}})$\;
\BlankLine

\tcp{Forward diffusion process}
$t \sim \mathcal{U}\{1, \ldots, T\}, \boldsymbol{\epsilon} \sim \mathcal{N}(\mathbf{0}, \mathbf{I}_d)$\;
$\mathbf{x}_t^{\text{norm}} \gets \sqrt{\bar{\alpha}_t} \, \mathbf{x}_{\text{gt}}^{\text{norm}} + \sqrt{1-\bar{\alpha}_t} \, \boldsymbol{\epsilon}$\;
\BlankLine

\tcp{Reverse diffusion}
$\hat{\boldsymbol{\epsilon}}_\theta \gets f_\theta(\mathbf{x}_t^{\text{norm}}, t; \mathcal{P}_{\text{obs}})$\;
$\hat{\mathbf{x}}_0^{\text{norm}} \gets \frac{1}{\sqrt{\bar{\alpha}_t}} \mathbf{x}_t^{\text{norm}} - \frac{\sqrt{1-\bar{\alpha}_t}}{\sqrt{\bar{\alpha}_t}} \hat{\boldsymbol{\epsilon}}_\theta$\;
$\hat{\mathbf{x}}_0 \gets \mathrm{Denormalize}(\hat{\mathbf{x}}_0^{\text{norm}})$\;
\BlankLine

\tcp{Denoising objective}
$\mathcal{L}_\epsilon \gets \|\boldsymbol{\epsilon} - \hat{\boldsymbol{\epsilon}}_\theta\|_1$\;
\BlankLine

\tcp{Hand-object penetration}
$\mathbf{d} \gets \text{SDF}(\mathcal{H}(\hat{\mathbf{x}}_0), \mathcal{P}_{\text{gt}})$\;
$\mathcal{L}_p \gets \sum_{i < |\mathcal{P}_{\text{gt}}|}\left[\max(0, \mathbf{d_i}) \cdot \exp\left(-\frac{t}{T/2}\right)\right]$\;
\BlankLine

\tcp{Contact Points Error}
$\hat{\mathbf{k}}, \mathbf{k}_{\text{gt}} \gets (\mathcal{K}(\hat{\mathbf{x}}_0), \mathcal{K}(\mathbf{x}_{\text{gt}}))$\;
$\mathcal{L}_k \gets \frac{1}{K}\sum_{j=1}^K \|\hat{\mathbf{k}}_j - \mathbf{k}_{\text{gt},j}\|_2$\;
\BlankLine
\Return{$\mathcal{L}_\epsilon + \lambda_p \cdot \mathcal{L}_p + \lambda_k \cdot \mathcal{L}_k$}\;
\end{algorithm}

\begin{figure*}[t!]
    \centering
    \includegraphics[width=1.0\linewidth]{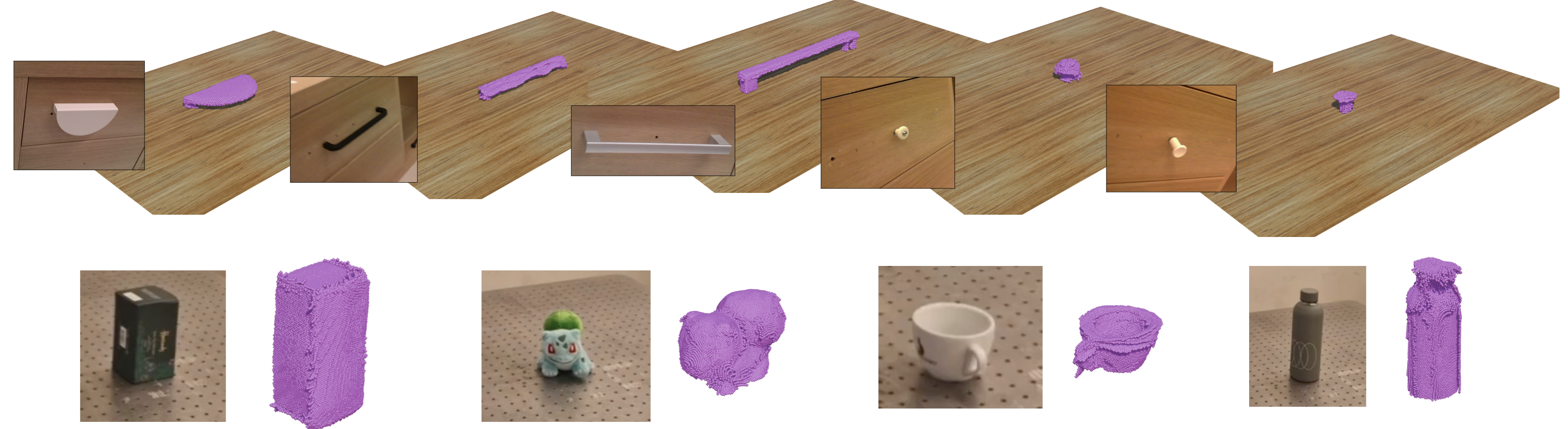}
\caption{\textbf{Real-world point cloud reconstructions.} We capture depth data for objects used in hardware experiments and reconstruct 3D point clouds through our multi-view fusion pipeline. Top row shows cabinet handles with their reconstructed point clouds. Bottom row shows real-world objects paired with their corresponding point cloud reconstructions. These point clouds serve as input to the diffusion model for grasp prediction.}

    \label{fig:real_pointclouds}
\end{figure*}
\parag{Training procedure.}
We train the diffusion model end-to-end using the AdamW optimizer with learning rate $\eta=2\times10^{-4}$ and default weight decay parameters ($\beta_1=0.9$, $\beta_2=0.999$, weight decay $w=0.01$).
Training is conducted for ${\sim}370{,}000$ steps with batch size 128.
We apply gradient clipping with maximum norm $\|\nabla\|_{\infty}=1.0$ to stabilize training dynamics. The whole training takes roughly 2 days on an RTX4090.
\\
\\
\parag{Diffusion noise schedule.}
We use $T=100$ diffusion steps during training with a linear $\beta$-schedule,
\begin{equation}
    \beta_t = \beta_{\min} + \frac{t-1}{T-1}(\beta_{\max} - \beta_{\min}), \quad t \in \{1,\ldots,T\},
\end{equation}
where $\beta_{\min}=0.0001$ and $\beta_{\max}=0.02$.

\parag{Inference.}
During sampling, we use the DDPM sampler with $T_{\text{sample}}=100$ denoising steps.
For real-world deployment, we generate $N_{\text{samples}}=16$ candidate grasps per object and select the first one that is feasible using the cuRobo path planner.
Optionally, we apply test-time guidance by using the gradient of the penetration depths of the point cloud with our hand model at the given sampling iteration.

\subsection{Hardware Deployment Details}
\label{sec:hardware}
\begin{figure}
    \centering
    \includegraphics[width=1.0\linewidth]{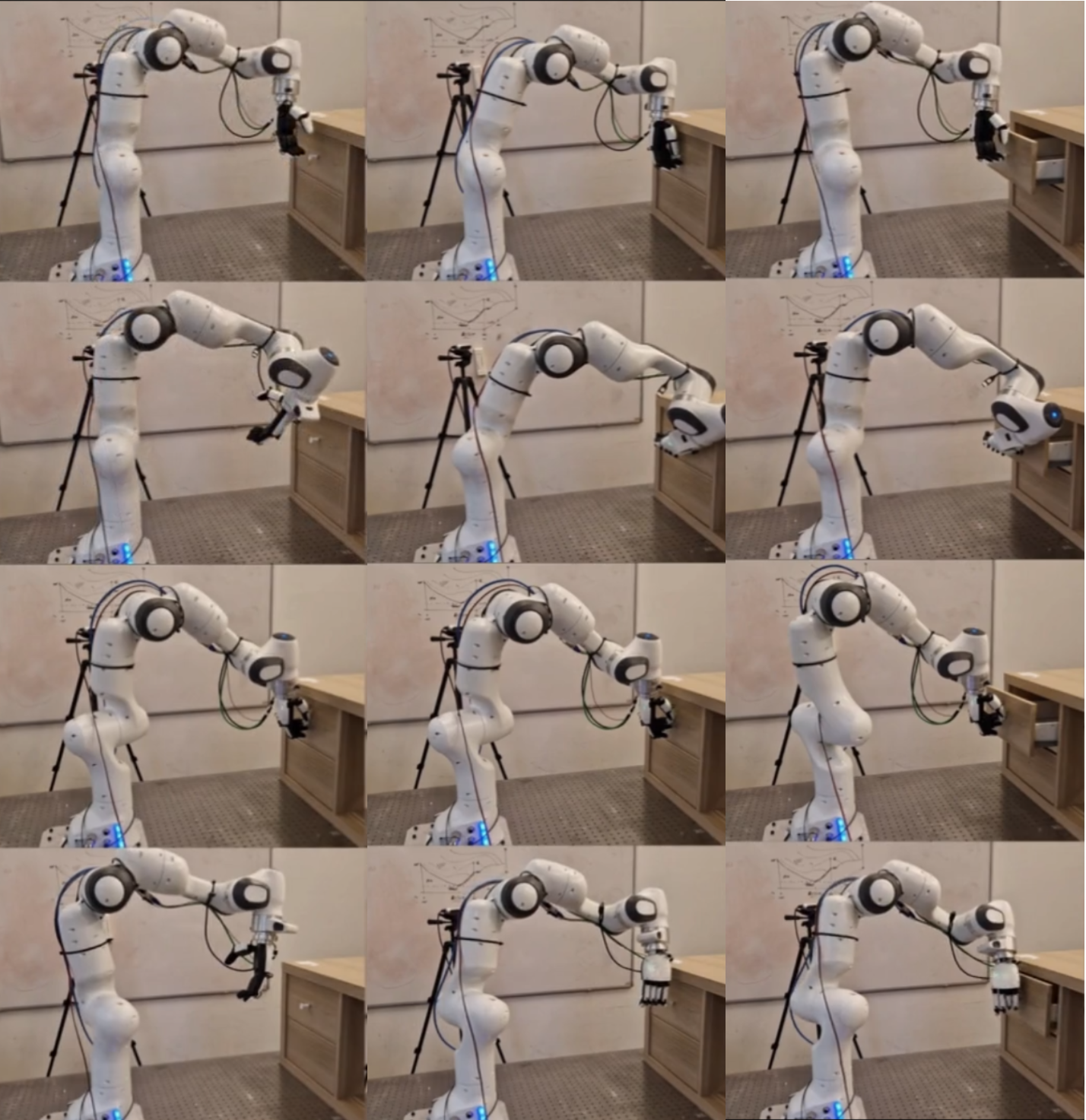}
    \caption{\textbf{Cabinet opening with knob-style handle.} Four successful grasp strategies for opening a cabinet with a \emph{knob-style handle}. Each row shows a different approach, with frames capturing the approach, grasp, and opening phases.}
    \label{fig:cabinet_knob1}
\end{figure}

\begin{figure}
    \centering
    \includegraphics[width=1.0\linewidth]{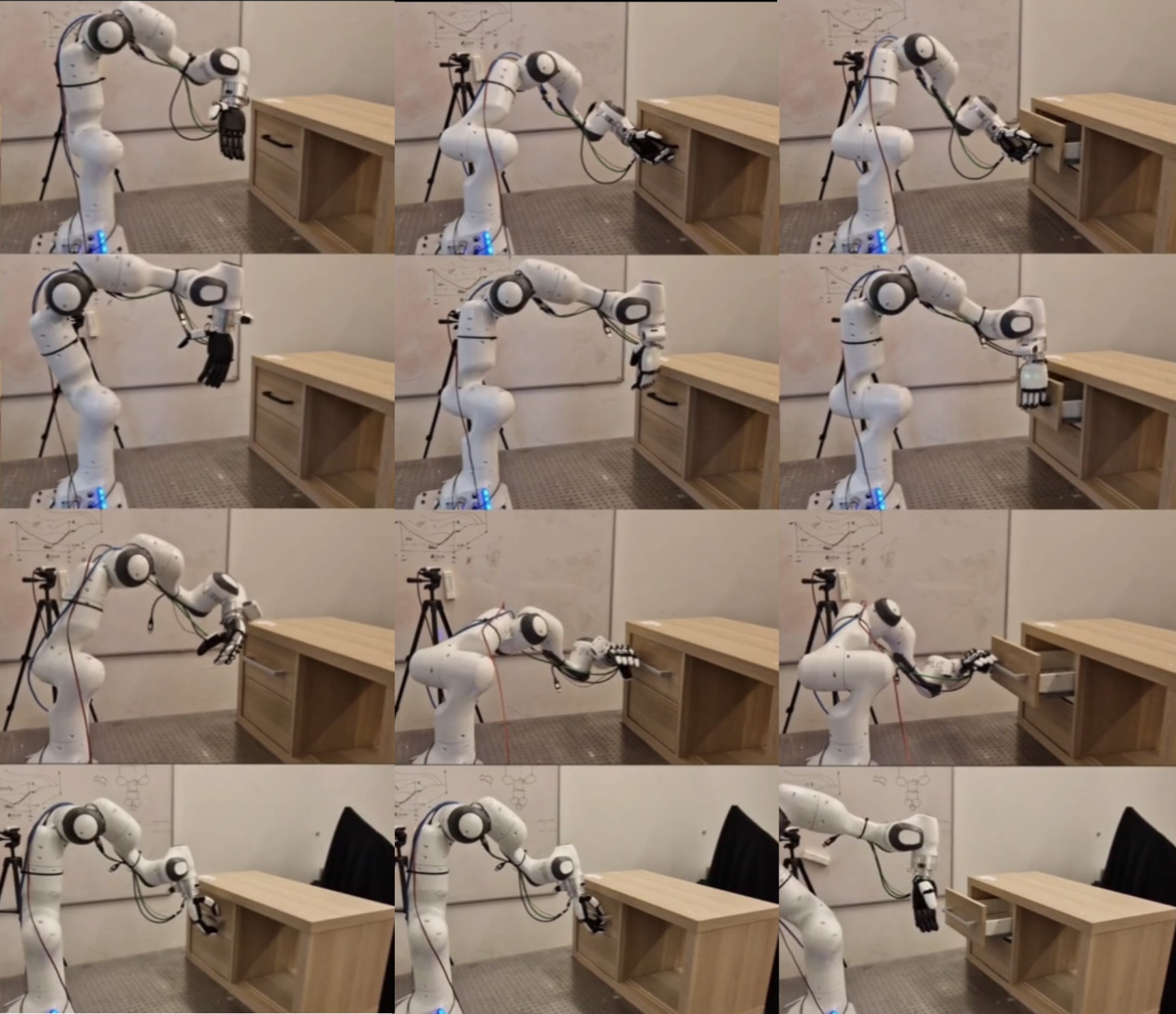}
    \caption{\textbf{Cabinet opening with bar-style handle.} Four successful grasp strategies for opening a cabinet with a \emph{bar-style handle}. Each row shows a different approach, with frames capturing the approach, grasp, and opening phases.}
    \label{fig:cabinet_bar}
\end{figure}

\begin{figure}
    \centering
    \includegraphics[width=1.0\linewidth]{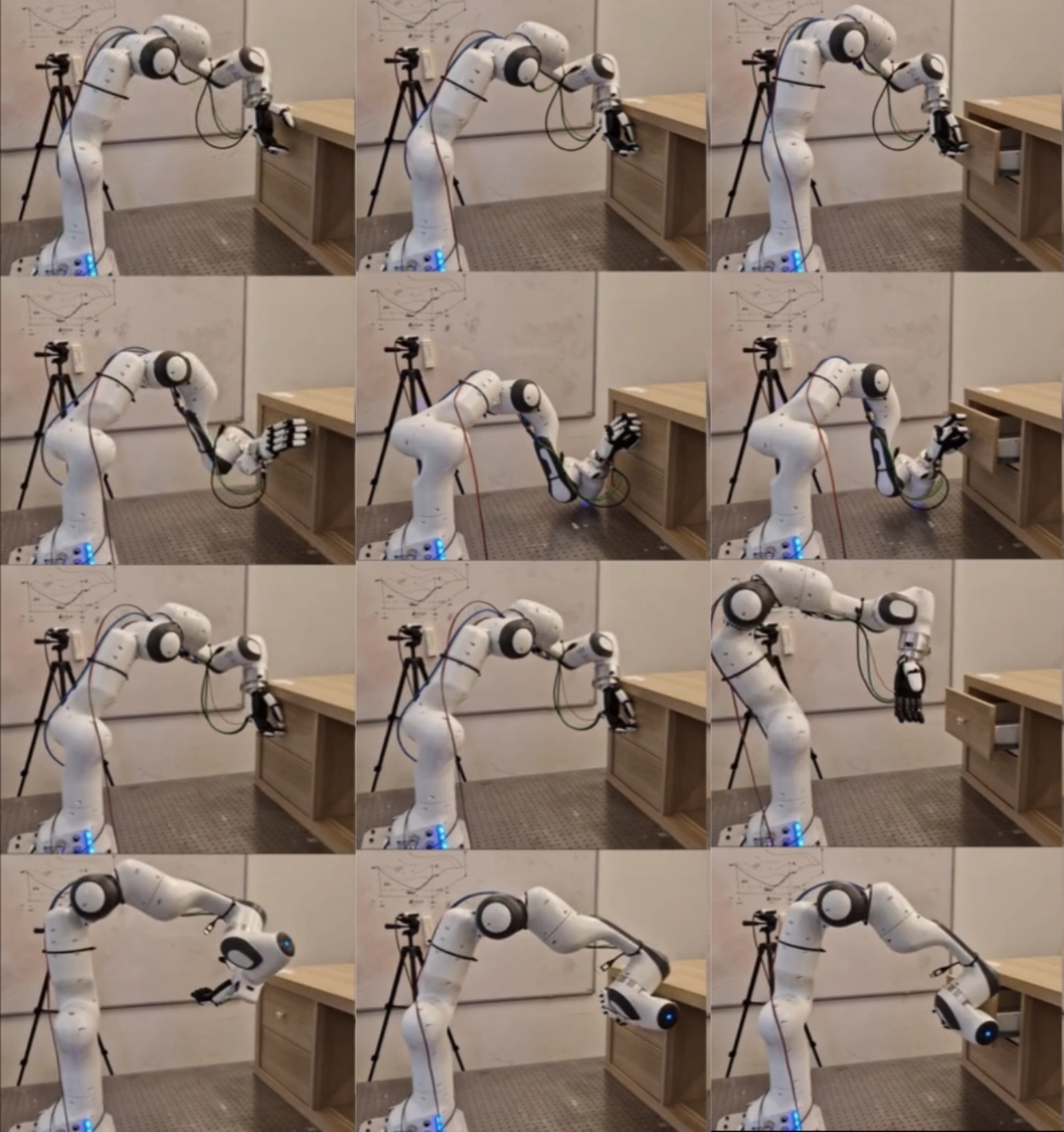}
    \caption{\textbf{Cabinet opening with knob-style handle (variant 2).} Four successful grasp strategies for opening a cabinet with a \emph{knob-style handle}. Each row shows a different approach, with frames capturing the approach, grasp, and opening phases.}
    \label{fig:cabinet_knob2}
\end{figure}

\begin{figure}
    \centering
    \includegraphics[width=1.0\linewidth]{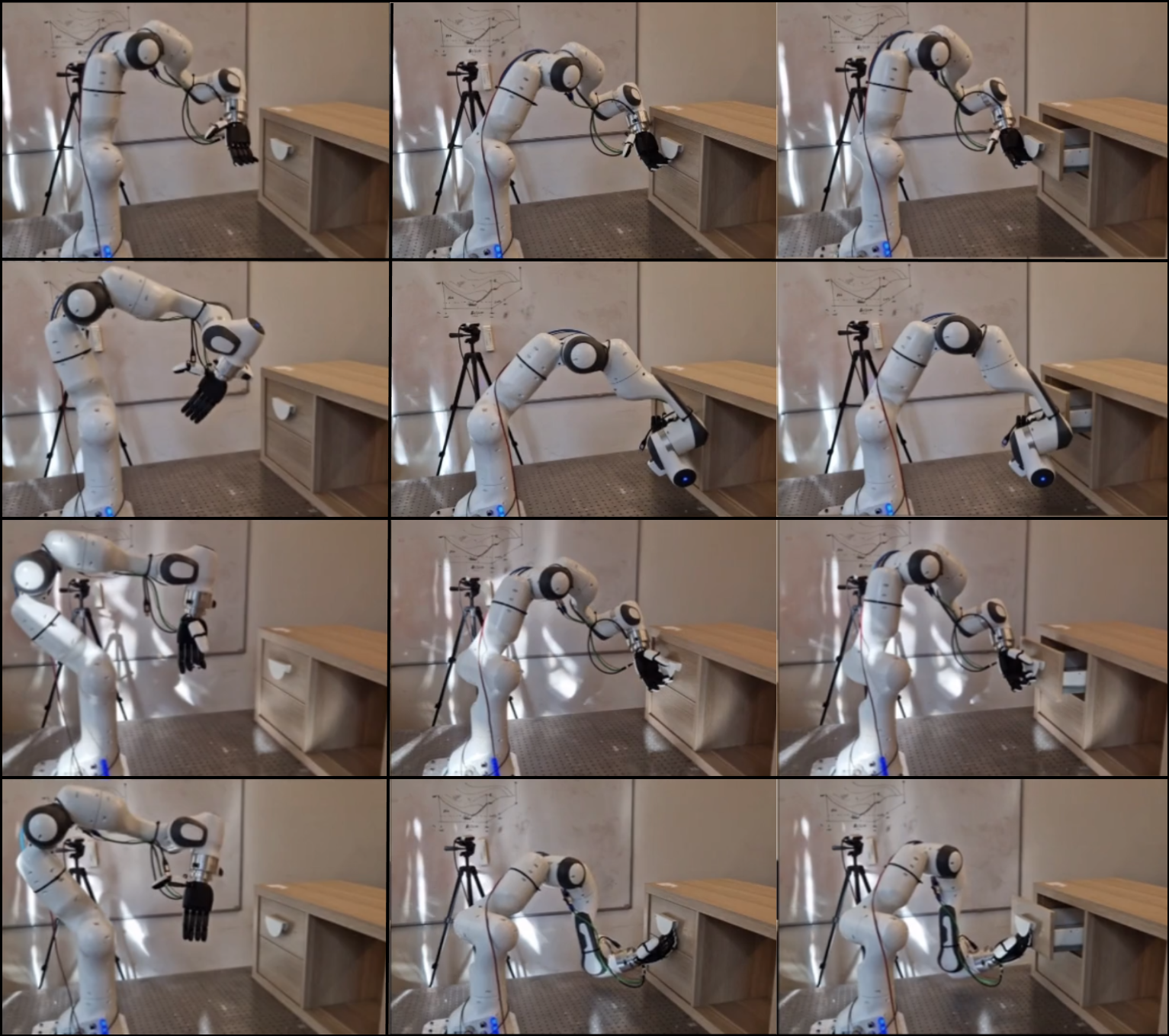}
    \caption{\textbf{Cabinet opening with shell-style handle.} Four successful grasp strategies for opening a cabinet with a \emph{shell-style handle}. Each row shows a different approach, with frames capturing the approach, grasp, and opening phases.}
    \label{fig:cabinet_shell}
\end{figure}

\begin{figure}
    \centering
    \includegraphics[width=1.0\linewidth]{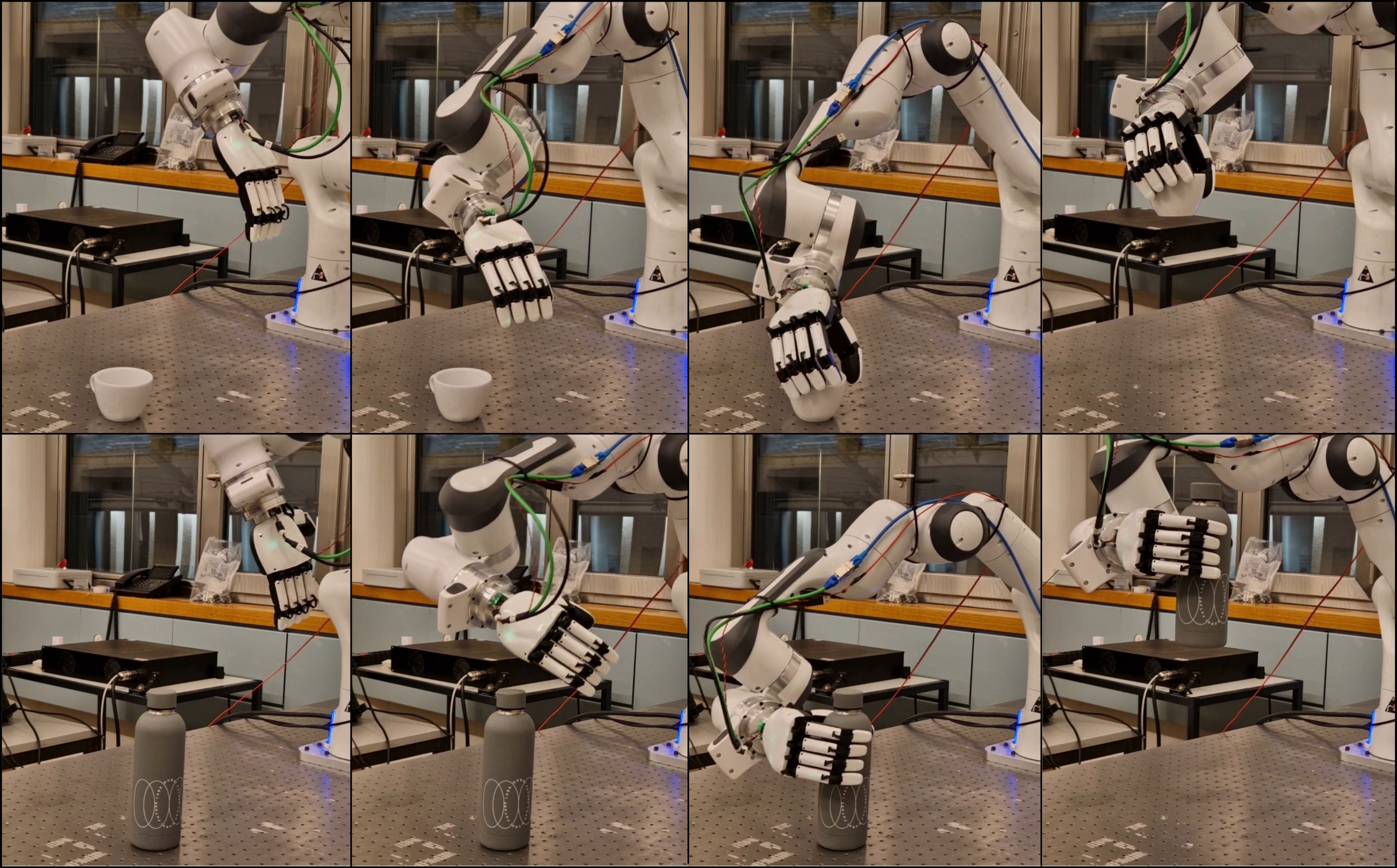}
 \caption{\textbf{Grasping household objects.} Successful grasp execution on a mug (top row) and bottle (bottom row). Frames show the approach, pre-grasp configuration, and final stable grasp for each object.}

    \label{fig:household_grasps}
\end{figure}

\begin{figure}
    \centering
    \includegraphics[width=1.0\linewidth]{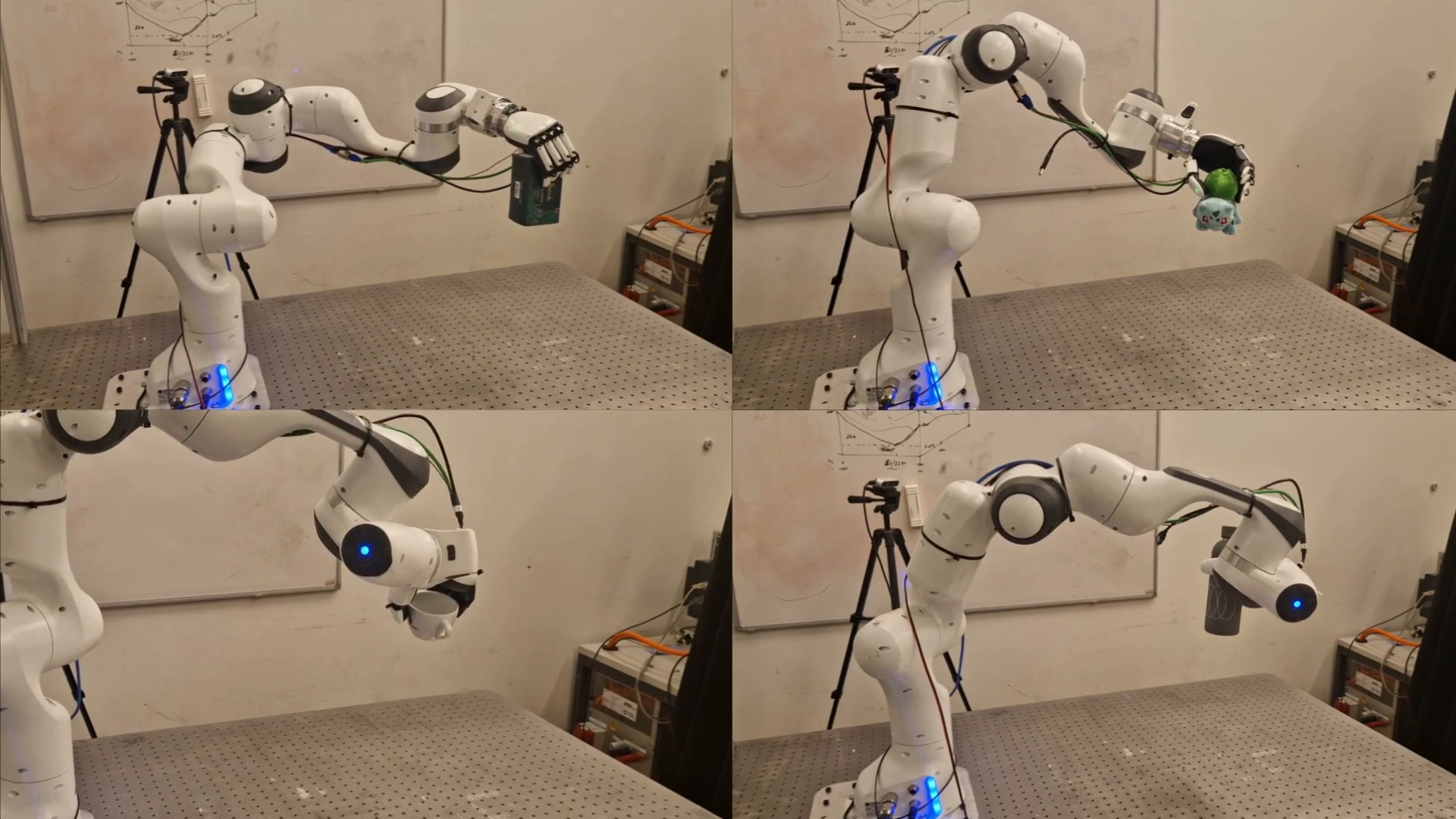}
    \caption{\textbf{Lifted object poses.} Successfully grasped and lifted household objects showing final stable configurations: tea box, toy figure, cup, and bottle (clockwise from top left).}

    \label{fig:lifted_objects}
\end{figure}

Figures~\ref{fig:cabinet_knob1},~\ref{fig:cabinet_bar},~\ref{fig:cabinet_knob2}, and~\ref{fig:cabinet_shell} show successful cabinet opening sequences with different handle types, demonstrating diverse grasp strategies. Figures~\ref{fig:household_grasps} and~\ref{fig:lifted_objects} illustrate successful grasping and manipulation of household objects.

\parag{Robot platform.}
We use a Franka Emika Panda 7-DoF robot arm equipped with the XHand dexterous hand.
The robot and hand are controlled using a joint impedance controller.

\parag{Perception system.}
For real-world experiments, we use an Intel RealSense L415 depth camera to capture RGB-D observations of target objects and handles. To address challenges inherent in single-frame depth acquisition, including sensor noise, occlusions, lack of texture, and limited coverage, we implement a multi-view reconstruction pipeline that fuses depth estimates from multiple viewpoints.

We scan each object by capturing RGB-D sequences from different viewpoints, equidistantly selecting $N_{\text{views}}=7$ frames. For each RGB frame, we apply monocular depth estimation using DepthAnything-v3~\cite{depthanything3}, which provides more reliable depth predictions than the RealSense sensor alone. Each view is then processed through a three-stage pipeline: (1) depth lifting converts the dense depth map into a 3D point cloud using camera intrinsics and estimated poses; (2) statistical outlier removal eliminates noise and spurious points; (3) normal estimation computes per-point normals oriented toward the camera.

After processing individual views, we fuse all $N_{\text{views}}$ point clouds into a unified coordinate frame and apply RANSAC-based plane segmentation to identify and remove the support surface. DBSCAN clustering then extracts the main object of interest. The resulting filtered point cloud $\mathcal{P}_{\text{obs}}$ is scaled to metric units using per-frame depth images to calculate the average scale factor, and serves as input to the diffusion model for grasp prediction.


Figure~\ref{fig:real_pointclouds} shows example reconstructions obtained with this pipeline for both cabinet handles and household objects used in our hardware experiments.

\begin{figure*}
    \centering
    \includegraphics[width=1.0\linewidth]{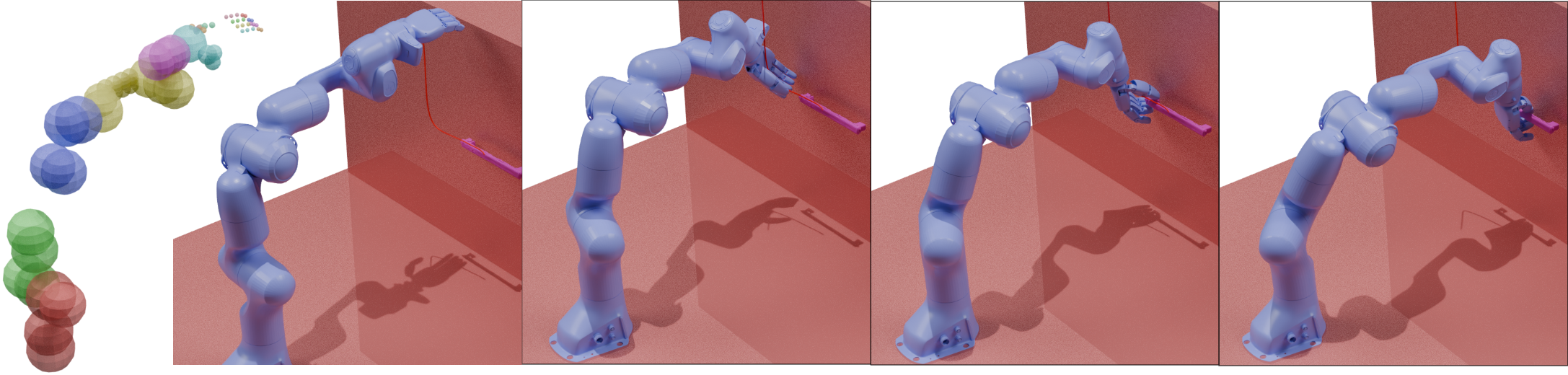}
     \caption{\textbf{Motion planning with cuRobo.} Collision-aware trajectory planning and execution for cabinet manipulation. The environment is approximated using simplified collision geometries: cubic representations for the floor and cabinet, and voxelized models for the handle point cloud. The sphere representation used for the robot is shown on the left. The sequence (left to right) shows successful approach and grasp execution on a bar-style handle.}

    \label{fig:curobo_planning}
\end{figure*}
\parag{Motion planning.}
We plan collision-free trajectories with cuRobo~\cite{sundaralingam2023curobo} using its MotionGen optimizer (Figure~\ref{fig:curobo_planning}).
Given a target end-effector pose $\chi_{\text{target}} \in SE(3)$ (and joint-state constraints for the hand $q_{\text{hand}}$, given by the predicted grasp configuration), we solve a trajectory optimization problem to find a collision-free path from the current configuration to the target.

For environment representation, we convert the reconstructed scene point cloud into a voxelized obstacle representation with voxel size $v=3\,\mathrm{mm}$ and cap the obstacle set at $N_{\max}=128$ voxels (uniform downsampling).
The robot is represented using sphere-based collision geometry with spheres covering the arm and fingers.
We model the table plane as a cube in the environment.
For the cabinet experiments, we additionally approximate the back panel of the drawer with a cube positioned at the support plane location.

Planning typically succeeds within $15\,\mathrm{s}$. If planning fails, we retry up to 5 times before selecting a new grasp that was precomputed by the diffusion model.

\subsection{Preference Alignment}
\label{sec:preference}

While stability is a necessary condition for successful grasps, it is not sufficient for generating human-preferred, natural-looking grasps.
We observe that stable grasps can vary significantly in their quality—some appear awkward, use sub-optimal contact regions, or exhibit unnatural hand configurations.
To address this, we collect human preference data and train a reward model to guide the evolutionary refinement toward grasps that are both stable and human-aligned.

\begin{figure*}
    \centering
    \includegraphics[width=1.0\linewidth]{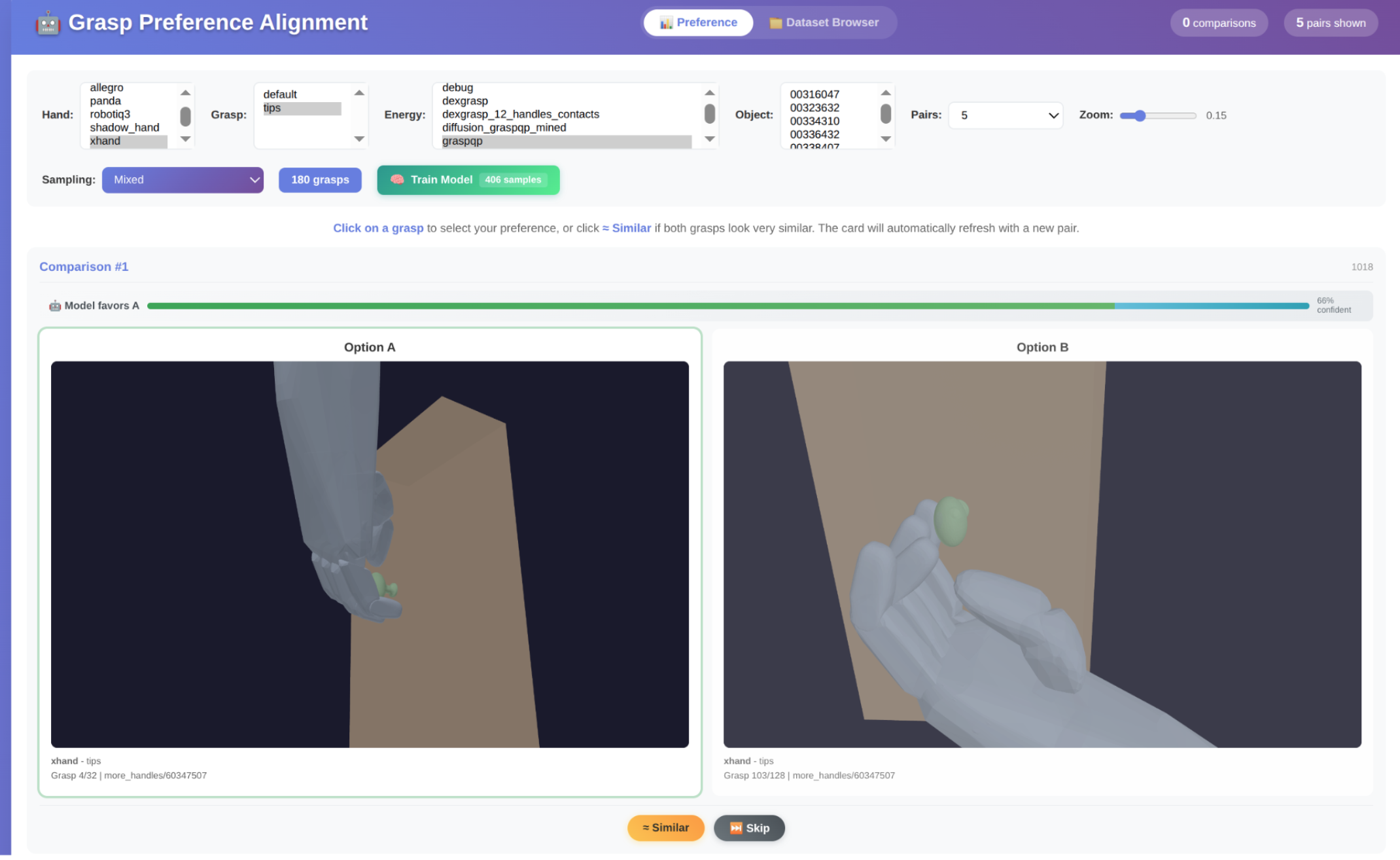}
    \caption{\textbf{Preference annotation interface.} The user interface used for collecting pairwise preference labels over grasp pairs. Annotators compare two grasps and select which appears more natural and human-intent-aligned, or mark them as similar.}
    \label{fig:annotation_ui}
\end{figure*}

\parag{Annotation protocol.}
We collect pairwise preference labels over grasp pairs $(G_a,G_b)$ using the web-based annotation interface shown in Figure~\ref{fig:annotation_ui}.
For each comparison, annotators view two rendered grasps from multiple viewpoints (front, side, top) showing the hand-object configuration.
Annotators are instructed to select the grasp that appears:
\begin{itemize}
\item More natural and human-like
\item Better positioned on the object (e.g., grasping handles at natural locations)
\item Using appropriate contact regions (e.g., fingertips vs. palm)
\item Less awkward or strained in hand configuration
\end{itemize}
Annotators can also select ``Similar'' if both grasps appear equally good or equally poor.
Each comparison takes a few seconds on average and a total of approximately 1{,}000 annotations are collected.

\parag{Reward model (PointNet++).}
We train a preference (reward) model $R_\psi(G,\mathcal{P})$ using a PointNet++-style set-abstraction network.
The model takes as input (i) an object point cloud with per-point features (6D, e.g., xyz+normals) and (ii) a set of hand keypoints with normals.
We concatenate these two point sets and add a binary ``type'' channel indicating object vs. hand points, resulting in 7D per-point features; PointNet++ uses the xyz coordinates as positions and the full 7D features as point attributes.
The network applies three set-abstraction layers (FPS + ball query + PointNetConv) and global max pooling, followed by an MLP head producing an embedding.
A final MLP maps the embedding to a scalar reward.

\parag{Training objective.}
We use a Bradley-Terry (logistic) preference loss. For a labeled comparison where $a$ is preferred over $b$, we minimize
\[
\mathcal{L}_{\text{pref}} = -\log \sigma\!\bigl(R_\psi(G_a,\mathcal{P}) - R_\psi(G_b,\mathcal{P})\bigr).
\]
For ``similar'' (tie) labels, we add a regularizer that encourages equal scores by penalizing the difference
\[
\mathcal{L}_{\text{sim}} = \mathbb{E}\!\left[(R_\psi(G_2,\mathcal{P}) - R_\psi(G_1,\mathcal{P}))^2\right].
\]
This term is weighted by a hyperparameter $\lambda_{equal}=0.5$ and scaled by the fraction of similar pairs in the minibatch.

\parag{Integration into refinement.}
During evolutionary refinement with preference guidance, we treat the predicted reward as an additional energy term $E_{\text{reward}}=R_\psi(G,\mathcal{P}) - b$ where $b=6.0$ is a bias term chosen as the $\min R_\psi(G,\mathcal{P})$ calculated over the training dataset. We add it to the fitness with weight $w_{\text{reward}}=10$.
The fitness becomes:
\[
\mathcal{F}(G) = w_{\text{lifetime}} E_{\text{lifetime}} - w_{\text{pen}} E_{\text{pen}} - w_{\text{dis}} E_{\text{dis}} + w_{\text{reward}} E_{\text{reward}}.
\]

\subsection{Reachability Analysis: 7-DoF vs.\ 6-DoF Robot Arms}
\label{sec:reachability_analysis}

\parag{Overview.}
To quantify the practical benefits of redundant kinematics for dexterous manipulation, we conduct a systematic reachability analysis comparing 7-DoF (Franka Panda) and 6-DoF (DynaArm) robot arms across a diverse set of handle-based manipulation scenarios. Figure~\ref{fig:reachability_analysis} presents this analysis, evaluating how many stable grasps remain kinematically reachable for different robot configurations and handle placements.

\parag{Experimental setup.}
We evaluate grasp candidates generated by two methods: (i) GraspQP (solid colors), our optimization-based baseline, and (ii) the evolutionary approach (hatched bars), using both 32 and 128 seed grasps. The evaluation uses 50 assets from the handles dataset, each placed at 12 different sampled positions in the robot's workspace. For each configuration, we test multiple object poses and compute inverse kinematics to determine which grasps are kinematically feasible for the robot arm.

\parag{Reachability metrics.}
We report three metrics that capture different aspects of grasp reachability:
\begin{itemize}
\item \textbf{Minimal Reachability} (Figure~\ref{fig:reachability_analysis}a): The mean of the minimum number of reachable grasps across all object poses. This metric represents the worst-case scenario, i.e., how many grasps remain available even when the object is in an unfavorable orientation. Higher values indicate more robust grasp coverage.
\item \textbf{Average Reachability} (Figure~\ref{fig:reachability_analysis}b): The mean number of reachable grasps averaged across all object poses. This metric reflects typical operating conditions and provides insight into the expected grasp diversity available during manipulation.
\item \textbf{Maximum Reachability} (Figure~\ref{fig:reachability_analysis}c): The mean of the maximum number of reachable grasps for the most favorable object pose. This metric shows the best-case grasp diversity and reveals the upper bound on manipulation options when the object orientation is advantageous.
\end{itemize}

\parag{Key findings and interpretation.}
The results demonstrate several important trends:
\noindent\textit{(1) 7-DoF advantage.}
The 7-DoF arm consistently achieves substantially higher reachability across all metrics. This kinematic redundancy enables reaching grasp poses infeasible with 6-DoF arms, particularly for handles positioned away from the optimal workspace.
\noindent\textit{(2) Impact of seed count.}
Increasing seeds from 32 to 128 improves reachability for both configurations, with larger marginal benefits for 7-DOF arms, suggesting redundant kinematics better exploit diverse grasp candidates.
\noindent\textit{(3) Evolutionary refinement.}
The evolutionary approach achieves comparable or better reachability than GraspQP, especially for average and maximum metrics, indicating that evolutionary refinement produces more kinematically accessible grasps and does not solely vary the finger joint angles.
\noindent\textit{(4) Worst-case robustness and deployment implications.}
Even in unfavorable orientations, 7-DOF arms with 128 seeds maintain 20+ reachable grasps versus <10 for 6-DOF arms. This motivates our 7-DOF Franka Panda platform choice: redundant kinematics provide critical advantages for handle manipulation in cluttered environments by maintaining diverse grasp options, enabling flexible deployment, and allowing nullspace motion for collision avoidance while maintaining end-effector poses.



\end{document}